%% file: main.tex
\documentclass[10pt,twocolumn,letterpaper]{article}

\usepackage{cvpr} 
\usepackage{amsfonts} 
\usepackage[accsupp]{axessibility}
\usepackage{dblfloatfix}
\usepackage{nicefrac}       
\usepackage{microtype}      
\usepackage{graphicx}
\usepackage{array}
\usepackage{amsmath}
\usepackage{amssymb}
\usepackage{pifont}
\usepackage{booktabs}
\usepackage{times}
\usepackage{graphicx}
\usepackage{amsmath}
\usepackage{amssymb}
\usepackage{dsfont}
\usepackage{siunitx}
\usepackage{multirow}
\usepackage{subcaption}
\usepackage{adjustbox}
\usepackage{enumitem}
\usepackage{tablefootnote}
\usepackage{bbm}
\usepackage{xspace}
\usepackage{mathtools}
\usepackage{color, colortbl}
\definecolor{Gray}{gray}{0.1}
\usepackage[normalem]{ulem}
\usepackage{tabularx}
\usepackage{wrapfig}
\usepackage{flushend}
\usepackage{longtable}
\usepackage{outlines}
\usepackage{bbding}
\usepackage[hang,flushmargin]{footmisc} 

\input{preamble}

\definecolor{cvprblue}{rgb}{0.21,0.49,0.74}
\definecolor{dgreen}{rgb}{0.0,0.6,0.0}
\usepackage[pagebackref,breaklinks,colorlinks,citecolor=cvprblue]{hyperref}

\def\paperID{4793} 
\def\confName{CVPR}
\def\confYear{2024}

\newcommand{\cmark}{\textcolor{dgreen}{\ding{51}}}
\newcommand{\xmark}{\textcolor{red}{\ding{55}}}

\newcommand{\dataset}{\texttt{EgoExoLearn}\xspace}
\newcommand\minisection[1]{\vspace{1mm}\noindent \textbf{#1}}

\title{\dataset: A Dataset for Bridging Asynchronous \\ Ego- and Exo-centric View of Procedural Activities in Real World}

\author{Yifei Huang$^{1\dagger\ddagger}$, Guo Chen$^{2\ddagger}$, Jilan Xu$^{3\ddagger}$, Mingfang Zhang$^{1\ddagger}$, Lijin Yang$^1$, Baoqi Pei$^4$ \\ Hongjie Zhang, Lu Dong$^5$, Yali Wang$^{6*}$, Limin Wang$^{2*}$, Yu Qiao$^*$ \\
OpenGVLab, Shanghai AI Laboratory \\ $^1$ The University of Tokyo \quad $^2$Nanjing University \quad $^3$Fudan University \\$^4$Zhejiang University \quad $^5$USTC \quad $^6$SIAT, CAS \\ 
}

\begin{document}


\maketitle
\let\thefootnote\relax\footnotetext{ $^*$Corresponding authors. $^\dagger$Project lead.  $^\ddagger$Equal key contributions.
}
\input{sec/0_abstract}    
\input{sec/1_intro}
\input{sec/2_related}
\input{sec/3_dataset}
\input{sec/4_benchmarks}
\input{sec/6_conclusion}
\input{sec/X_suppl}
{
    \small
    \bibliographystyle{ieeenat_fullname}
    \bibliography{main}
}


\end{document}

%% file: preamble.tex
\usepackage[dvipsnames]{xcolor}
\usepackage[inkscapelatex=false]{svg}


%% file: sec/0_abstract.tex
\begin{abstract}
\vspace{-1em}
Being able to map the activities of others into one's own point of view is a fundamental human skill even from a very early age. 
Taking a step toward understanding this human ability, we introduce \dataset, a large-scale dataset that emulates the human demonstration following process, in which individuals record egocentric videos as they execute tasks guided by exocentric-view demonstration videos.
Focusing on the potential applications in daily assistance and professional support,
\dataset contains egocentric and demonstration video data spanning 120 hours captured in daily life scenarios and specialized laboratories. 
Along with the videos we record high-quality gaze data and provide detailed multimodal annotations, formulating a playground for modeling the human ability to bridge asynchronous procedural actions from different viewpoints.
To this end, we present benchmarks such as cross-view association, cross-view action planning, and cross-view referenced skill assessment, along with detailed analysis. We expect \dataset can serve as an important resource for bridging the actions across views, thus paving the way for creating AI agents capable of seamlessly learning by observing humans in the real world. The dataset and benchmark codes are available at \url{https://github.com/OpenGVLab/EgoExoLearn}.
\end{abstract}

%% file: sec/1_intro.tex
\vspace{-1em}
\section{Introduction}
\label{sec:intro}

\begin{figure*}
    \centering
	\vspace{-1.5em}
	\includegraphics[width=\linewidth]{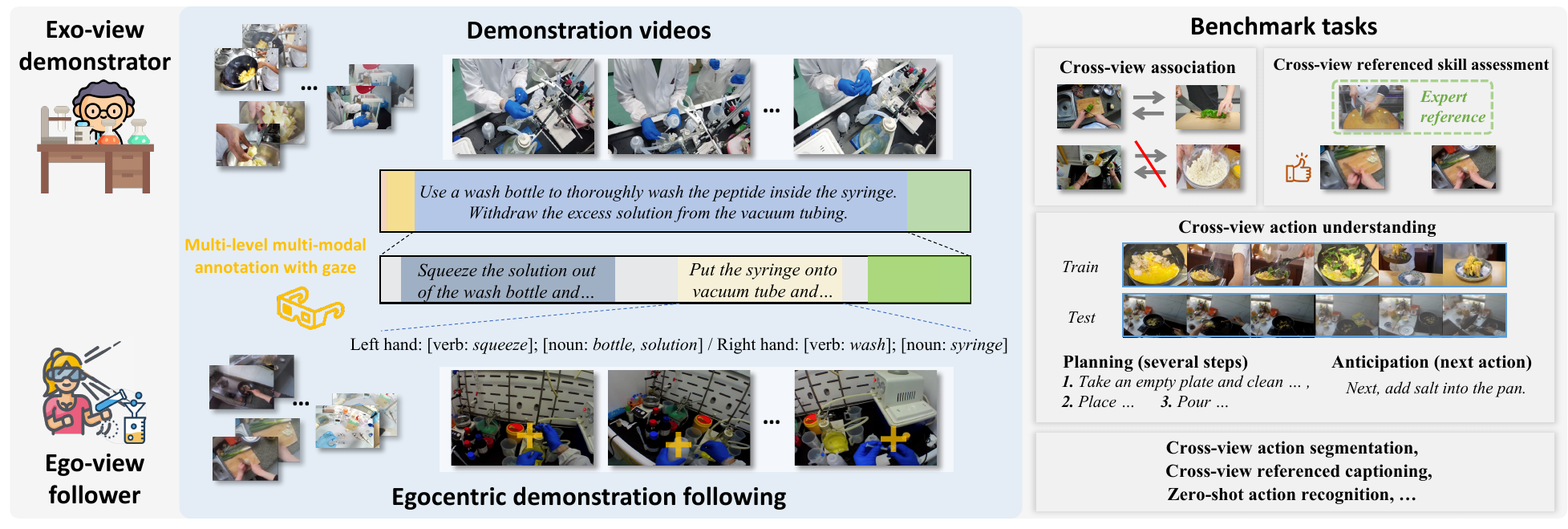}
	\vspace{-1.5em}
	\caption{\dataset emulates the human asynchronous demonstration following process. It contains demonstration videos of multiple tasks, together with egocentric videos recorded by participants replicating the procedure after watching the demonstrations. The dataset comprises gaze signals and fine-grained multi-level multi-modal annotations, enabling the exploration of key features in this context such as cross-view association and cross-view action planning. }
 \vspace{-1em}
	\label{fig:teaser}
\end{figure*}

Even as a child, humans can observe the actions of others and then map them to their own view~\cite{ramsey2021watch,bandura2008observational,schaal1996learning,hodges2007modelled}. With this ability to asynchronously bridge activities from egocentric and exocentric views~\cite{premack1978does,rizzolatti2004mirror}, humans can watch others' demonstrations and replicate the procedures in a new environment. This ability is especially beneficial when actual physical trials carry the potential of high costs~\cite{fryling2011understanding}, \textit{e.g.}, conducting dangerous chemical experiments. 

In the wake of recent advancements in AI systems, one goal for the next generation of AI agents is to perform tasks in a more embodied setting~\cite{plizzari2023outlook}. 
However, different from humans, training these AI agents usually requires demonstration videos taken in a similar environment~\cite{wang2023mimicplay,liu2023egocentric} and from a congruent perspective with the AI agents, (\textit{e.g.}, the egocentric point of view~\cite{lee2017learning,jang2022bc,ALFRED20,wong2022assistq}). While great effort has been made into the collection of egocentric data in different scenarios~\cite{ego4d,epickitchen,assembly101}, it remains crucial for the AI agents to directly learn from demonstration videos taken in a different place and from a different viewpoint~\cite{yu2018one,hua2021learning}. 
Realizing this capability can unleash the full potential of public instructional video data~\cite{howto100m} and is also useful in the human-robot cooperation scenario, especially in novel environments~\cite{liu2020skill,lallee2010human,wang2020see}.

Current works towards this goal can be roughly divided into two directions. One way is to learn models in simulated environments~\cite{habitat19iccv,pauly2021o2a,li2021meta,brockman2016openai,pan2023copilot}, but it remains difficult for models in this setting to generalize in the real world~\cite{HoloAssist2023}. The other direction is to learn from human activity in real-world scenarios. However, attempts to directly combine existing multiview datasets often yield datasets of inferior quality or scale~\cite{xue2023learning,wang2023learning}. Meanwhile, the few existing datasets in this direction~\cite{charadesego,rai2021home,assembly101} only record ego- and exo-view videos in the same environment and in a time-synchronized manner. 
In reality, when following demonstrations it is often needed to bridge a series of procedural actions performed in a different place and at a different time.
However, currently, no dataset is available for the exploration of how to bridge asynchronous procedural activities in realistic egocentric and exocentric viewpoints.

To address this lack of dataset issue, we introduce \dataset, a large-scale dataset containing demonstration videos and corresponding egocentric videos where the camera wearers follow the demonstrations and perform the same task in a different environment, as shown in Figure~\ref{fig:teaser}. Targeting two potential applications, \textit{i.e.}, daily assistance and professional support, \dataset consists of 747 video sequences spanning a total of 120 hours of footage, ranging from daily food-making to specialized laboratory experiments. 
Notably, the egocentric videos in \dataset contain eye gaze signals showing humans' visual attention while performing the task. This provides a valuable cue for better bridging the actions in ego- and exo-viewpoints.

We take one more step forward by analyzing human ability in bridging asynchronous ego- and exo-view actions and, accordingly, introduce new tasks and benchmarks that we believe can form building blocks for the development of next-stage embodied AI agents with similar abilities. 
When humans perform an action, he/she can associate and describe the undergoing action in the egocentric view with the corresponding action in the demonstration. With the knowledge from demonstration videos, humans can know the needed action steps and predict what the next steps should be. 
Besides, through the comparison with the demonstration, humans can also assess their level of skills. 

Based on the above analysis, we design benchmarks of 1) cross-view association, 2) cross-view action understanding, 3) cross-view referenced skill assessment, and 4) cross-view referenced video captioning.
Each benchmark is meticulously defined, annotated, and supported by baseline implementations. In addition, we pioneeringly explore the role of gaze in these tasks.
We hope our dataset can provide resources for future work for bridging asynchronous procedural actions in ego- and exo-centric perspectives, thereby inspiring the design of AI agents adept at learning from real-world human demonstrations and mapping the procedural actions into robot-centric views. 

%% file: sec/2_related.tex
\section{Related Work}
\label{sec:related}

\minisection{Ego-exo datasets.}
While there exist works that associate existing datasets to explore how activities can be bridged between them, these associated datasets are often limited in scale~\cite{chen2019ta3n,yingzhen2018disentangled,finebio} or quality~\cite{wang2023learning}, meanwhile focusing only on single actions captured from the same view~\cite{pan2020adversarial,materzynska2019jester,wei2023unsupervised}.
As for actions from different views, apart from multi-view fixed camera datasets~\cite{ben2021ikea,kuehne2014language,li2021igibson,Corona_2021_WACV}, there also exist datasets with both ego- exo-centric view videos~\cite{charadesego, assembly101,ego4d,rai2021home,egoexo4d}. These datasets are either recorded in the same environment~\cite{charadesego,rai2021home,jia2020lemma} or record time-synced multi-view videos in the same environment with primary focuses on pose/activity understanding grounded in the 3D world~\cite{kwon2021h2o, egobody, assembly101}. Our dataset offers a more challenging and realistic scenario, where egocentric camera wearers learn to complete the tasks demonstrated by exocentric demonstration videos. This setting complements these datasets by focusing on high-level procedural actions.

The only dataset conceptually similar to ours is the recently proposed AE2 dataset~\cite{xue2023learning}, where the goal is to learn view-invariant representation from unpaired ego and exo videos. This dataset combines ego and exo videos from five public datasets~\cite{epickitchen,de2009guide,kwon2021h2o,kuehne2011hmdb,zhang2013actemes} and a newly collected ego tennis forehand dataset. However, due to the difficulty in associating existing ego-exo datasets, the AE2 dataset is relatively small where the largest subset contains only 322 clips. Also, this dataset only focuses on clip-level actions, and thus cannot feature the real-world demonstration following setting, which usually requires multimodal, task-centric procedural knowledge. Instead, our \dataset is much larger in scale (100x more clips), while offering gaze and fine-grained multimodal annotations facilitating multi-faceted analysis of ego-exo action understanding. 

\minisection{Egocentric video datasets.} 
In line with the recent development in wearable cameras~\cite{Aria2023}, multiple egocentric video datasets~\cite{damen2018scaling, ego4d, hoi4d, ragusa2021meccano, EgoProceLECCV2022,VISOR2022,jia2022egotaskqa,yu2023fine,song2024ego4d} have been proposed. 
Different from previous egocentric datasets, the egocentric videos in  \dataset feature a demonstration-following setting. We believe \dataset provides a playground for developing tools to bridge asynchronous procedural activities from ego- and exocentric viewpoints.
The setting of our \dataset complements existing datasets like Ego4D and can benefit from their rich knowledge and representations.

\minisection{Egocentric gaze.}
Gaze can indicate visual attention and contains valuable information about human intent~\cite{huang2015using,zhang2022can}, thus is used in a diverse range of areas such as human-computer interaction~\cite{hutchinson1989human,zander2010combining,kim2019eyes}, and augmented reality~\cite{rivu2020stare,park2008wearable}. In computer vision, efforts have been made to leverage gaze in various tasks~\cite{li2013learning,li2018eye,huang2020mutual,li2015delving,min2021integrating,huang2018predicting,lai2022eye,xu2015gaze,razin2017learning,huang2020ego}. However, with the previous absence of large-scale egocentric datasets that include gaze, this avenue of research is currently under-explored~\cite{plizzari2023outlook,li2021eye, egobody, zheng2022gimo, jang2019epic}. Our \dataset offers calibrated gaze positions for all egocentric videos. Thanks to our unique setting, our dataset enables the integration of gaze in egocentric video understanding and the exploration of the role of gaze in the cross-view context.

\minisection{Egocentric and ego-exo video understanding.} The unique recording perspective of egocentric videos presents a series of challenges including but not limited to action understanding~\cite{kazakos2019epic,furnari2019would,girdhar2021anticipative,sudhakaran2019lsta,wang2021interactive,wang2023ego,plizzari2022e2,wang2023memory,radevski2023multimodal,huang2022compound,chen2024video}, hand detection~\cite{cai2016understanding,shan2020understanding,zhang2022fine,goyal2022human}, and video-language understanding~\cite{kang2021video,huang2023weakly}.
These form fundamental building block techniques of embodied AI~\cite{nagarajan2021shaping}, VR/AR~\cite{jiang2021egocentric,bettadapura2015egocentric,wang2023scene,liu2022joint}, and human-robot interaction~\cite{marina2022head,xin2023learning,nagarajan2020ego,li2019deep}. Since most egocentric datasets are smaller in scale compared with general datasets which contain mostly exocentric view videos~\cite{frozenintime,xue2022advancing,zellers2021merlot}, it is possible to leverage exocentric video data to improve model performance on egocentric videos~\cite{weinland2010making}. There are typically three main directions: joint view-invariant learning~\cite{xue2022advancing,wang2023learning,xu2024retrieval}, domain adaptation~\cite{xia2022incomplete}, and knowledge distillation~\cite{li2021ego}. In this work, we evaluate all these directions in our benchmarks. 

%% file: sec/3_dataset.tex
\section{Dataset}

\subsection{Data Collection}
\label{sec:data_collection_scenario}

\minisection{Scenarios and tasks.} We consider procedural goal-oriented tasks ranging from daily food-making to specialized laboratory-based experiments. This selection is grounded in their exemplification of two prospective areas where future embodied AI agents would need the ability to bridge ego-exo activities: daily-life assistance and professional support. Specifically, \dataset incorporates 5 types of daily tasks (\textit{e.g.}, cooking) and 3 types of specialized laboratory tasks (\textit{e.g.}, solid-phase peptide synthesis). We record egocentric videos in 4 different kitchens and 3 different labs.
Other details are provided in the supplementary. 

\minisection{Data collection procedure.} Before the start of each collection session, participants are required to complete a questionnaire gathering basic demographic information and their self-evaluated expertise in executing the designated task. This questionnaire also highlights the ethical, privacy, and security considerations. Then in each session, participants will be asked to choose one or several exocentric view demonstration videos from a provided list and carefully learn the detailed procedures. Once they feel ready, they will wear Pupil Invisible Glasses~\cite{pupil}, complete the gaze calibration, and begin to replicate the task performed in the demonstration videos. While not encouraged, participants are permitted to revisit the demonstration video during the recording.

After each recording session, the participants are asked to re-do the gaze calibration to ensure gaze data fidelity. 
For the 5 daily tasks, the exocentric demonstration videos are manually curated from online video platforms such as YouTube. For the lab experiments, the exocentric demonstration videos are tutorials recorded by senior lab members. 


\begin{figure}
    \centering
    \includegraphics[width=\linewidth]{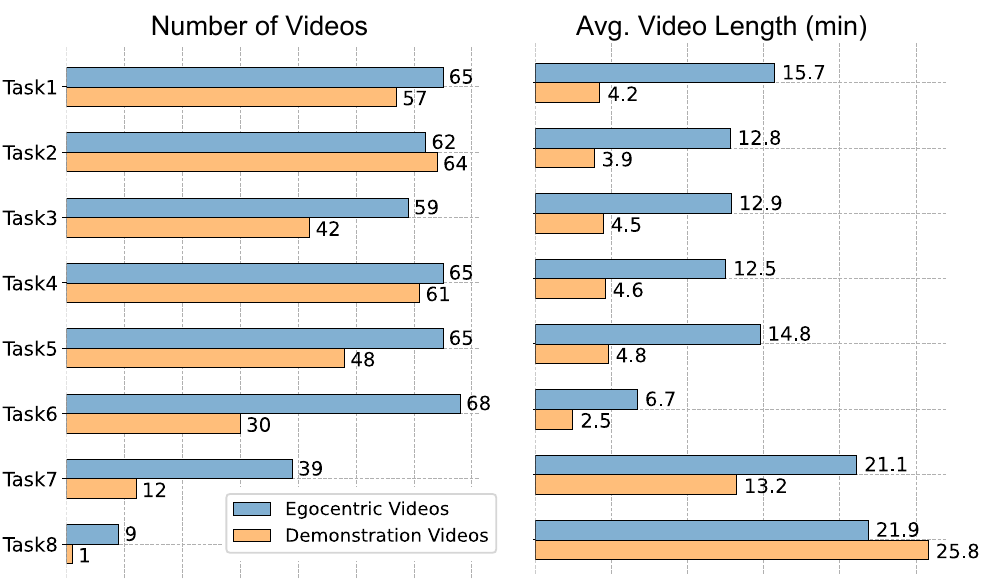}
    \caption{The number of videos per task (left) and the average duration of each video per task (right). Task1 to Task5 represent the 5 daily tasks and the remaining are three tasks in specialized laboratories. In each recording session of the egocentric video, one participant may learn from multiple demonstration videos and one demonstration video may be watched by several participants.}
    \label{fig:numlen}
    \vspace{-1em}
\end{figure}

Figure~\ref{fig:numlen} shows the distribution of the 120 hours of data. Since most demonstration videos are meticulously edited to remove repeated steps, the average length of demonstration videos is lower than the egocentric videos which record the full procedure. As a result, the \dataset contains 432 egocentric videos totaling 96.5 hours and 315 demonstration videos spanning 23.5 hours. This difference in video length poses a unique challenge when bridging ego- and exo-view activities for future research endeavors.

\begin{figure*}
    \centering
    \includegraphics[width=\linewidth]{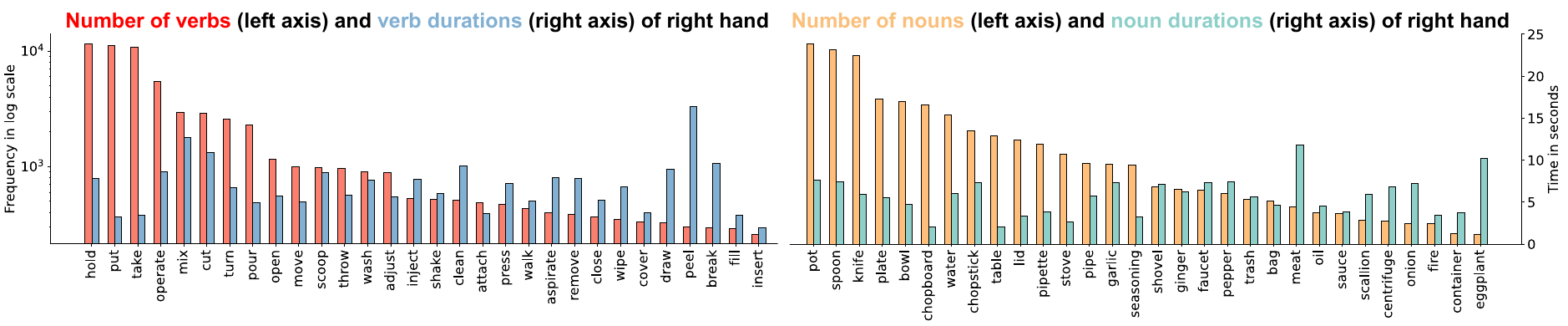}
    \vspace{-2em}
    \caption{Occurrence and duration distribution of the annotated fine-level verbs and nouns associated with the right hand.}
    \vspace{-1em}
    \label{fig:verbnoun}
\end{figure*}

\subsection{Annotation}
\label{sec:annotation}
To facilitate our dataset in the development of algorithms that can effectively bridge the gap between ego and exo viewpoints, we provide detailed multi-modal human annotations. Our pipeline of annotation contains four stages detailed in the following paragraphs. Each of these stages is subject to a rigorous manual quality check involving no fewer than two individuals for verification and validation.

\minisection{Coarse-level language annotation.}  In this step, we ask annotators to annotate the coarse actions in the videos. Like the previous works~\cite{assembly101,HoloAssist2023}, the coarse actions are defined as a middle-level step for accomplishing a task and can be divided into multiple fine actions. For instance, ``Prepare the pork" in the task of twice-cooked pork, and ``Suction filtration" in solid-phase peptide synthesis.
Three types of annotation are given in this step: 1) the temporal interval, consisting of start and end timestamps; 2) the action label; and 3) a language description of the video within the annotated interval. 
We specifically request annotators to focus on elucidating ``what is done," ``how it is done," and ``the purpose of this step" in their language descriptions. We define a total of 39 categories of coarse-level actions in this stage and acquire 41.2 coarse-level annotations per video with an average length of 21.5 seconds.

\minisection{Fine-level language annotation.} Based on the coarse-level annotations, in this step we request the annotators to provide annotations for the fine-level actions. The fine-level actions are the atomic actions like ``take knife" or ``pull syringe plunger". Unlike the first step, annotators are instructed to furnish language descriptions that specifically emphasize ``which hand is used", 
 ``what object is used" and ``why it is used".
For the first two steps, we employ a two-round manual annotation checking to ensure the annotation quality.

\minisection{Translation \& parsing.} To ensure linguistic precision, all the annotators give the language description using their native language~\cite{epickitchen}. For non-English annotations, we employ ChatGPT and Google Translation API to translate them into English. Subsequently, for the fine-level annotations, we employ specific rules and utilize tools such as NLTK~\cite{bird2006nltk} and Spacy~\cite{vasiliev2020natural} to
extract the verbs and nouns associated with each segment. During this stage, we confine the selection of verbs and nouns to the predefined taxonomy offered by Ego4D~\cite{ego4d}, while also manually introducing supplementary verbs and nouns that are absent in the taxonomy. Since our manual annotation specifies the engagement of the left/right hand, we can extract multiple verbs and nouns for each segment, meanwhile attributing them to the respective involvement of the left or right hand. After manual checking, we obtain a total of 95 verb and 254 noun categories in the fine-level annotation. In Figure~\ref{fig:verbnoun} we show the occurrence of the top 30 categories of verbs and nouns attributed to the right hand. More statistics can be found in the supplementary.
\input{tab/dataset_comparison}
\input{tab/annotation_comparison}

\minisection{Skill level annotation.} Since self-assessed skill level is not perfectly suitable for skill assessment, we identify several representative skills and assign human annotators to assess their skills. The annotation follows a pairwise ranking scheme, where annotators are presented with pairs of videos of the same action, and instructed to determine which video demonstrates a higher skill level. We prepare 40,191 video pairs and ensure that 4 different annotators annotate each pair. After filtering out pairs with less than 3 consistent opinions, we get a collection of 34,239 valid video pairs.

\subsection{Statistics \& Comparisons}  
To the best of our knowledge, there is no dataset that follows the same setting as ours for a direct comparison. Therefore, we enumerate various aspects of our dataset and conduct a comparative analysis with relevant datasets in Tables \ref{tab:dataset_comparison} and \ref{tab:annotation_comparison}. \dataset distinctively enriches the domain with its ``visual demonstration following" setting. Beyond this unique setting, it stands as the first egocentric dataset that includes temporal bounded language captions, annotated cross-view associations, and multi-label video segments. 

%% file: tab/dataset_comparison.tex
\begin{table*}[t]
\centering
\scriptsize
\setlength{\tabcolsep}{7.5pt}
\def\arraystretch{1}
\begin{tabular}{@{}l|ccccc|cccccc@{}}
\toprule
Dataset & Settings  & \begin{tabular}[c]{@{}c@{}}Unique \\ Hours \end{tabular} & \begin{tabular}[c]{@{}c@{}}Ego \\ +Exo?\end{tabular} & \begin{tabular}[c]{@{}c@{}}Instruction \\ following?\end{tabular} & \begin{tabular}[c]{@{}c@{}}Visual\\ instruction?\end{tabular} & Gaze & \begin{tabular}[c]{@{}c@{}}Coarse\\ Action\end{tabular} & \begin{tabular}[c]{@{}c@{}}Fine\\ Action\end{tabular} & \begin{tabular}[c]{@{}c@{}}Dense \\ Caption\end{tabular} & Association & Skill \\ \midrule
Meccano~\cite{ragusa2021meccano}  & Industry & 7 &   \xmark   &   \cmark   &  \xmark   &    \xmark   &      \xmark     &      \cmark    &   \xmark      & \xmark  & \xmark \\
EGTEA~\cite{li2018eye}  & Cooking & 28 &   \xmark   &   \cmark   &  \xmark   &    \cmark   &      \xmark     &      \cmark    &   \xmark      & \xmark  & \xmark \\
EK-100~\cite{epickitchen}   & Cooking  & 100 &   \xmark   &       \xmark     &  \xmark   &    \xmark   &      \xmark     &      \cmark    &   \xmark      & \xmark  & \xmark \\
HoloAssist~\cite{HoloAssist2023}  & Assistive & 166 &   \xmark   &   \cmark   &  \xmark   &    \cmark   &      \cmark     &      \cmark    &   \xmark      & \xmark  & \cmark \\
Ego4D~\cite{ego4d}  & Multiple & 3670 &   \xmark   &   \xmark   &  \xmark   &    $\diamond$   &      $\diamond$     &      $\diamond$    &   $\diamond$    & \xmark  & \xmark \\
\midrule
H2O~\cite{kwon2021h2o}  & Desk & 1 &   \cmark   &   \xmark   &  \xmark   &    \xmark   &      \xmark     &      \cmark    &   \xmark      & \xmark  & \xmark \\
LEMMA~\cite{jia2020lemma}  & Daily & 10 &   \cmark   &   \xmark   &  \xmark   &    \xmark   &      \xmark     &      \cmark    &   \xmark      & \xmark  & \xmark \\
HOMEAGE~\cite{rai2021home}  & Daily & 25 &   \cmark   &   \xmark   &  \xmark   &    \xmark   &      \cmark     &      \cmark    &   \xmark      & \xmark  & \xmark \\
CharadesEgo~\cite{charadesego}  & Daily & 34 &   \cmark   &   \xmark   &  \xmark   &    \xmark   &      \xmark     &      \cmark    &   \xmark      & \xmark  & \xmark \\
Assembly101~\cite{kwon2021h2o}  & Desk & 42 &   \cmark   &   \cmark   &  \xmark   &    \xmark   &      \cmark     &      \cmark    &   \xmark      & \xmark  & \xmark \\
\midrule
\dataset (ours)  & Daily \& Lab & 120 &   \cmark   &   \cmark   &  \cmark   &    \cmark   &      \cmark     &      \cmark    &   \cmark      & \cmark  & \cmark \\
\bottomrule
\end{tabular}
\caption{\textbf{Comparison to related datasets on settings (left) and annotations (right).}  ``Unique hours" refers to the cumulative duration of distinct video recordings, counting only one camera's footage for simultaneous recordings of the same activity. $\diamond$: Partially included. }
\vspace{-1em}
\label{tab:dataset_comparison}
\end{table*}

%% file: tab/annotation_comparison.tex
\begin{table}[t]
\centering
\adjustbox{width=\linewidth}{
\setlength{\tabcolsep}{3pt}
\def\arraystretch{1.2}
\begin{tabular}{lcc|cc|ccccc}
\hline
Dataset & \begin{tabular}[c]{@{}c@{}}\#\\ videos\end{tabular} & \begin{tabular}[c]{@{}c@{}}Avg. \\  min\end{tabular} & \begin{tabular}[c]{@{}c@{}}\# \\ segs\end{tabular} & \begin{tabular}[c]{@{}c@{}}Avg.\\  sec\end{tabular} & \begin{tabular}[c]{@{}c@{}}verb \\ classes\end{tabular} & \begin{tabular}[c]{@{}c@{}}noun \\ classes\end{tabular} & \begin{tabular}[c]{@{}c@{}}\# verb\\ / seg\end{tabular} & \begin{tabular}[c]{@{}c@{}}\#noun \\ / seg\end{tabular} & \begin{tabular}[c]{@{}c@{}}\#words\\ / seg\end{tabular} \\ \hline
CharadesEgo~\cite{charadesego} & 7860 & 0.5 & 69k & 1.8 & 32 & 37 & 1 & 1 & 0 \\
HOMEAGE~\cite{rai2021home} & 5700 & 0.9 & 26k & - & 29 & 86 & - & - & 0 \\
EK-100~\cite{epickitchen} & 700 & 8.5 & 90k & 3.1 & 97 & 300 & 1 & 1.2 & 3.0 \\
Assembly101~\cite{assembly101} & 4321 & 7.1 & 83k & 1.7 & 24 & 90 & 1 & 1 & 0 \\ 
Ego4D~\cite{ego4d} & 991 & 26.4 & 77k & 8.0 & 115 & 478 & 1 & 1 & 7.4 \\ \hline
Ours-ego & 432 & 13.4 & 64k & 4.6 & 95 & 254 & 1.8 & 2.5 & 16.9 \\
Ours-exo & 315 & 4.5 & 14k & 4.7 & 82 & 251 & 2.3 & 3.0 & 19.4 \\ \hline
\end{tabular}
}
\caption{\textbf{Contemporary egocentric datasets.} We show only the fine-level actions for a fair comparison. For Ego4D~\cite{ego4d}, we select the closest subtask of “forecasting” following~\cite{assembly101}.
}
\vspace{-0.5em}
\label{tab:annotation_comparison}
\end{table}

%% file: sec/4_benchmarks.tex
\section{Dataset Properties \& Benchmarks}
\subsection{Dataset Properties}

\dataset stands out from current egocentric and ego-exo datasets due to several unique properties.

\minisection{Ego-Exo demonstration following setting.} The most distinguished property of \dataset is the ego-exo demonstration following context. Egocentric video recorders are instructed to follow the steps in exocentric demonstration videos to perform the same task but in a different environment. 
This setting closely emulates the human observational learning process~\cite{hodges2007modelled,bandura2008observational} and can be instrumental in designing embodied AI agents that learn from alternative perspectives while executing tasks from their own viewpoint. 

\minisection{Fine-grained vision-language annotations with gaze.} To facilitate a deeper analysis, we equip our dataset with rich, multimodal, fine-grained annotations. 
\dataset is the first egocentric dataset featuring high-quality captions, evidenced by the number of words per segment in Tab.~\ref{tab:annotation_comparison}. Different from Ego4D where captions are associated with only single timestamps, our captions come with manually annotated start and end timestamps. For better visual perception across ego-exo views, we give verb and noun labels associated with the specific hand. 
One aspect from which we can analyze the human's ability to bridge ego-exo activities is through gaze. \dataset is also augmented with calibrated eye-gaze signals.
These annotations enable the understanding of human ability to bridge ego-exo activities from diverse perspectives, which we posit will benefit the next-generation embodied AI agents~\cite{li2023behavior,deitke2020robothor}.

\subsection{Benchmarks}
To evaluate the ability of bridging asynchronous ego-exo procedural activities, we introduce 4 new benchmarks: 1) cross-view association, 2) cross-view action understanding, 3) cross-view referenced skill assessment, and 4) cross-view referenced captioning. The cross-view action understanding benchmark is further subdivided into three subtasks: cross-view action anticipation, cross-view action planning, and cross-view action segmentation. Additionally, we explore the role of gaze in assisting these tasks. We also benchmark models on zero-shot and supervised fine-grained action recognition tasks for reference, following~\cite{epickitchen,HoloAssist2023}. Note that we carefully split our dataset to eliminate annotation leak across benchmarks. Due to the space limit, we only provide partial content of definition, annotation, results, and analysis, leaving more complete details in the supplementary.

\begin{figure*}
    \centering
    \includegraphics[width=\linewidth]{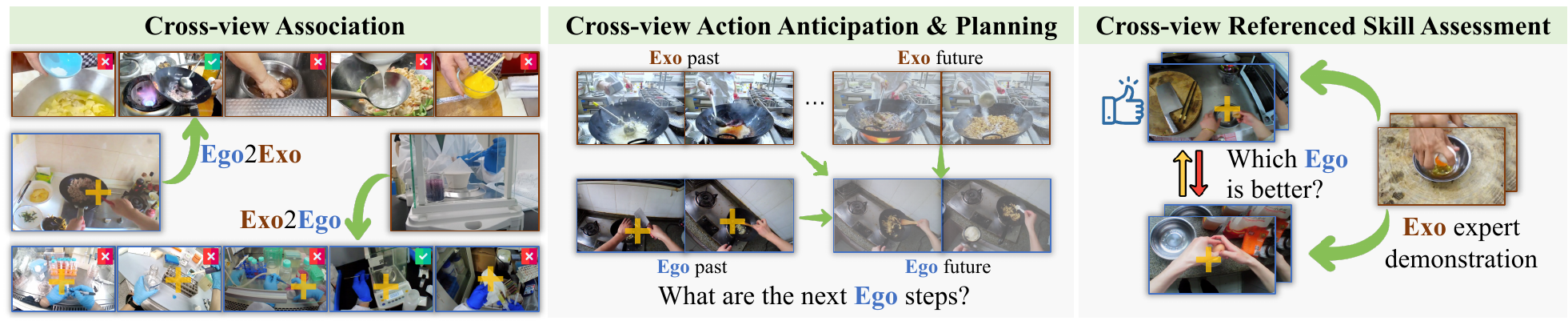}
    \vspace{-2em}
    \caption{Concept of the 3 benchmarks of cross-view association (Sec.~\ref{benchmark:association}), cross-view action anticipation \& planning (Sec.~\ref{benchmark:action}) and cross-view reference skill assessment (Sec.\ref{benchmark:skill}) in this section. Other benchmarks can be found in the supplementary material.}
    \vspace{-1em}
    \label{fig:enter-label}
\end{figure*}

\subsubsection{Cross-view association} 
\label{benchmark:association}
\minisection{Motivation.} 
One straightforward indicator of the ego-exo activity bridging is the ability to associate the same semantics across ego- and exo-views. This benchmark focuses on equipping models with this cross-view association ability. An application of this ability is assistants in AR that can show expert demonstration videos when the human is confused~\cite{zhang2023x}. Another potential application is the embodied AI agent that can explain its decision~\cite{yang2024video,wallkotter2021explainable,wang2023generating}.

\minisection{Problem settings.}
We formulate this association benchmark as a cross-view multiple-choice association problem. Specifically, we consider two different cross-view association settings: ego2exo and exo2ego. In the case of ego2exo, given an egocentric video, the model is asked to predict the corresponding exocentric video performing the same action from a candidate choice set of exocentric samples, and vice versa for the exo2ego setting. 
For both ego2exo and exo2ego settings, we use 20 candidate samples for each query.
The evaluation metric is the averaged Top-1 accuracy.

\minisection{Annotations.} We meticulously construct the ground-truth ego-exo pairs via a semantic-aware matching process. It is composed of five stages with details in the supplementary material: 
(1) Scenario Matching. 
(2) Noun and Verb Matching. 
(3) Sentence Matching with LLM. 
(4) Negative Sampling. 
(5) Two-round Manual Verification. 
Notably, we do not provide such pairs for the training set and leave the modeling of cross-view association on unpaired samples to be further explored for the community.

\minisection{Baseline model.} 
We adopt three types of baseline models, \textit{i.e}., ego-only, exo-only, and ego-exo, which refer to the data we used during training. Under all the settings, we leverage the paired video and caption and jointly train a video encoder and a text encoder using the contrastive loss~\cite{clip}. We use TimeSformer-B~\cite{timesformer} as the video encoder and clip-text~\cite{clip} as the text encoder. We initialize both encoders using the EgoVLP~\cite{egovlp} pre-trained weights. During testing, we obtain the ego/exocentric video representations using the video encoder. The prediction is defined as the one with the highest normalized cross-view video feature similarity among all candidates.
On top of these models, we evaluate the effectiveness of gaze in associating egocentric video and exocentric video. This is achieved by replacing the original egocentric video with spatially cropped videos using the gaze positions as the cropping center.

\minisection{Related work.} 
Prior works on pre-trained vision-language models for multiple choice association/questioning ~\cite{videoclip,clipbert,simvlm,allinone,egovlp,lavila} are generally pre-trained on either exocentric~\cite{frozenintime,howto100m} or egocentric~\cite{ego4d} datasets, only revealing weak cross-view bridging ability. Another line of works explores cross-view learning~\cite{wu2013cross,ardeshir2018exocentric,xu2018joint,ho2018summarizing} by either transferring the knowledge from one view to the other~\cite{li2021ego} or training view-invariant video understanding models~\cite{wang2023learning,xue2023learning}. Different from previous works, our cross-view association benchmark is in a more realistic but challenging setting, by evaluating the model's ability to associate asynchronous activities across ego- and exo-views.

\minisection{Experiment results.}
We first evaluate several vision-language models via zero-shot transfer. These models are pre-trained either on egocentric videos, {\em i.e.} EgoVLP~\cite{egovlp} and LaViLa~\cite{lavila}, or exocentric videos, {\em i.e.,} InternVideo~\cite{internvideo}. As shown in Tab.~\ref{tab:association}, without using gaze, EgoVLP generally outperforms the others on both validation and test sets. By introducing gaze information, InternVideo receives a decent improvement, especially on Exo2Ego. 

As for fine-tuned models, Tab.~\ref{tab:association}, reveals that models trained solely on single-view data struggle with cross-view association. Training ego-only models using gaze-cropped egocentric videos results in substantial improvements, significantly outperforming those trained on center-cropped videos. This again highlights the importance of gaze in enhancing cross-view association. Based on our observation, the regions around gaze help the cluttered ego videos become visually similar to exo videos where the primary object is salient. Last, we show the baseline result when co-trained on both egocentric and exocentric videos. The model in this setting shows the strongest cross-view association ability over single-view models. Findings from this benchmark underline the limitation of current models in associating activities across ego and exo views, and point towards the potential benefits of integrating gaze into the association.
\input{tab/retrieval}

\subsubsection{Cross-view action anticipation \& planning}
\label{benchmark:action}
\minisection{Motivation.} The procedural actions are not guaranteed to be identical between the two views due to practical constraints. Thus, we design benchmarks for cross-view action anticipation and planning to enable a thorough understanding and transfer of necessary steps (or actions) for task completion, bridging the gap between views and considering real-world conditions. A practical application of this benchmark can be seen in human-robot collaboration scenarios. For instance, an embodied AI agent, after observing a human perform the first part of a task, could effectively take over and complete the remaining half of the task based on the particular environmental situation.

\minisection{Problem settings.} 
For both the cross-view anticipation and cross-view planning, the goal is to anticipate the future activities in one view, given labeled training data only in another view. For the cross-view anticipation task, we focus on predicting the verb and noun categories of the next fine-level action $\tau=1$ second into the future. 
Given the multilabel nature of our verb and noun annotations in each fine-level segment, we perform multiclass anticipation. Performance is evaluated by class-mean Top-5 recall as per~\cite{epickitchen}. For the cross-view planning task, we aim to generate the next $K=8$ steps of coarse-level actions. We adopt ED@$K$ as the evaluation metric following the setting of Ego4D LTA~\cite{ego4d}.

\minisection{Annotations.} 
For action anticipation, we use the fine-level verb and noun annotations, and then take their intersection between ego and exo videos to constrain them in the same closed set. 
To evaluate the model more effectively, we further control the long-tail degree of the data by and filter out the tail categories that occur less than \nicefrac{1}{100} of the highest occurrence category.
For the action planning task, we directly adopt the coarse-level action annotations and regard the start timestamp of each segment as one action step. More details can be found in the supplementary material.

\input{tab/cross_action_under}

\minisection{Baseline model.}
We explore three distinct directions to implement cross-view baseline models.
The first direction is based on unsupervised domain adaptation (UDA), treating one view as the source domain and the other as the target domain. 
This method operates within an unsupervised training framework, using labels from the source domain and video data from both domains~\cite{da2022dual,wei2023unsupervised,sahoo2021contrast,song2021spatio,kim2021learning,munro2020multi,yang2022interact}. 
We adopt CLIP~\cite{clip} + TA3N~\cite{chen2019ta3n} as the baseline model.
The second direction entails knowledge distillation (KD)~\cite{hinton2015hinton-distillation,pan2020spatio,girdhar2019distinit}, allowing the model trained on one view to learn knowledge of the other view, under the assumption that a teacher model of the other view is available. We equip CLIP with a distillation approach based on~\cite{li2021ego} to transfer knowledge from the teacher model to a newly created student model.
The third and most straightforward direction is co-training (CT) using the data from both views to encourage the model to discover correlations between them directly. For using gaze, we also crop the video based on the gaze positions.

We comprehensively consider four evaluation settings. The ``Ego-only" and ``Exo-only" settings do not involve cross-view understanding, thus we use zero-shot evaluation serving as the references. The Ego2Exo and Exo2Ego settings are the cross-view settings. For UDA, ``Ego2Exo'' is defined as utilizing the egocentric view as the source domain and the exocentric view as the target domain. In the context of KD, ``Ego2Exo'' indicates we initially train a teacher model on egocentric data, followed by the training and distillation of the student model on exocentric data. For CT, we merge both egocentric and exocentric datasets through direct concatenation. We report results on the test set and put the validation set results in the supplementary.

\minisection{Related work.}
Prior works~\cite{assembly101,kuehne2014language,fu2010multi,vyas2020multi} on multi-view action understanding mainly focus on synchronized multi-view videos. Some work studied transferring knowledge~\cite{li2021ego} from one view to the other or training the view-invariant~\cite{wang2023learning,xue2023learning,zheng2013learning} video models.
Our cross-view benchmarks seek to evaluate the ability of models to bridge asynchronous actions across views, which is more challenging yet realistic.

\minisection{Experiment results.} 
Table~\ref{tab:sta-and-lta} presents the results of action anticipation and planning on the test set.
The first block of results shows a significant performance gap when models trained exclusively on one view are tested on the other view. This underscores the inherent differences in activities captured in the two views. Remarkably, even without relying on any specific cross-view method, the inclusion of gaze information markedly diminishes the disparity between egocentric and exocentric data. 
Leveraging techniques such as UDA or KD we can see improved performance compared with direct zero-shot inference in the first block. 
Since CT can utilize both egocentric and exocentric data, it achieves the best performance in all setups.
Moreover, from the comparison in all settings using or not using gaze, it is clear that gaze serves as a surprisingly effective signal to mitigate the gap between activities in the two views, although naively designed.
These results take the first step in the potential directions for better bridging the cross-view activities, setting a foundation for future advancements in the field of cross-view action understanding.

\subsubsection{Cross-view referenced skill assessment}
\label{benchmark:skill}
\minisection{Motivation and problem setting.} 
We propose a novel task of cross-view referenced skill assessment leveraging the unique setting and annotations of our dataset. This task goes beyond the traditional pairwise ranking often used in skill assessment~\cite{doughty2018s,lam2022machine} by incorporating an expert demonstration video as a reference point. This demonstration provides a model of the ideal execution of an action, offering a standard against which to compare skill levels. In this task, the input is a pair of egocentric video clips $C_{ego1},C_{ego2}$ of the same action and an exo-view demonstration video clip $C_{exo}$ as the reference, and the output is a choice $c \in \{ego1,ego2\}$ of the egocentric clip that demonstrates a higher skill level. Moreover, we can also assess the skill level by the relation between action and gaze. 
This benchmark evaluates a model's ability to bridge asynchronous ego-exo dynamics. A practical application is an AR system that helps humans in skill acquisition by providing targeted feedback based on skill level assessment and expert demonstration.

\minisection{Annotations.} 
For the cross-view referenced skill assessment benchmark, we concentrate on four types of representative actions, as detailed in Tab.~\ref{tab:skill_assess}. 
We use the skill level annotations in Sec.~\ref{sec:annotation} in this benchmark. For each pair of videos, the annotation is provided by four distinct annotators to minimize subjective bias. We ensure credibility by checking the transitivity of annotations and removing the pairs with less than 3 agreements among 4 annotators.
We then append an exo-view demonstration video clip of the same action, forming video triplets as the model input. 

\minisection{Baseline model.} 
Our baseline model is built upon a pairwise ranking skill assessment model RAAN \cite{doughty2019pros}. Without loss of generality, assuming $ego1$ is the video showing a higher skill level, we apply the following approaches to leverage the reference exo-view demonstration video: 1) Triplet loss (TL). The feature distance between $C_{exo}$ and $C_{ego1}$ should be closer to the distance between $C_{exo}$ and $C_{ego2}$. 2) Relation network (RN). Inspired by \cite{sung2018learning}, we employ a relation network that concatenates the features of the ego and exo clips. This network is designed to discern which of the two egocentric video clips bears a closer relation to the demonstration video in terms of skill level. The way to gaze is consistent with the other benchmarks.

\minisection{Related work.} 
Several previous works on skill assessment aim to directly regress a score based on professional ratings~\cite{parmar2019and, ahmidi2017dataset, yu2021group, tang2020uncertainty, liu2021towards,li2022surgical}. We adopt a more general approach of pairwise ranking since no absolute score is available in most real-world skills~\cite{bertasius2017baller,li2019manipulation}. Previous works in this direction only use a pair of videos that are in either egocentric view~\cite{doughty2018s} or in exo view \cite{doughty2019pros,li2019manipulation}. Differently from their works, we explore the cross-view activity bridging ability by examining how demonstration videos from exo view can benefit skill assessment in ego views. 

\input{tab/skill_assessment}

\minisection{Experiment results.} We first evaluate the performance of previous works under the conventional pairwise setting. In the upper block of Table~\ref{tab:skill_assess}, both methods \cite{doughty2018s,doughty2019pros} receive a clear performance boost when gaze is used. This aligns with behavioral science findings that experts and novices have different gaze patterns in the same task~\cite{liu2009Who,wilson2010Psychomotor}. With the exocentric demonstration as reference, both the relation network method (RN) and the triplet loss method (TL) can leverage the reference video and bring performance improvement. Further adding gaze we can get the best performance. 
However, the marginal improvement observed with the inclusion of the exocentric reference suggests that current models may still struggle to fully bridge asynchronous activities across egocentric and exocentric views. There remains ample room for new improvement and innovation.

%% file: tab/retrieval.tex
    

\begin{table}[t]
\centering
\resizebox{\columnwidth}{!}{
  \begin{tabular}[t]{lccccc}
    \toprule
     \multirow{2}{*}{Method}& \multirow{2}{*}{Gaze} & \multicolumn{2}{c}{Val}  & \multicolumn{2}{c}{Test} \\
     \cmidrule(r){3-4}  \cmidrule(r){5-6}
     & & Ego2Exo & Exo2Ego & Ego2Exo & Exo2Ego \\
    \midrule
    
    \emph{Zero-shot} \\
    Random  &\XSolidBrush & 12.7 & 15.0	&	14.1	&13.4 
  \\
    EgoVLP~\cite{egovlp} &\XSolidBrush & 28.8 & 27.2  &	32.1&	28.9 \\
    LaViLa~\cite{lavila}  &\XSolidBrush &	22.6 &	24.9	&28.7	&25.7\\
    InternVideo~\cite{internvideo} &\XSolidBrush & 27.0 & 21.2 &	30.6&	21.7\\
    EgoVLP~\cite{egovlp} &\Checkmark & 28.8 &	29.7  & 31.5 & 28.9 \\
    LaViLa~\cite{lavila}  &\Checkmark &	21.9 &	21.4 & 30.3 & 25.9 \\
    InternVideo~\cite{internvideo} &\Checkmark & \textbf{30.9} & \textbf{32.3} &\textbf{33.3}& \textbf{32.2}  \\
    \midrule
    \emph{Fine-tuned} \\
    Exo-only &\XSolidBrush & 	42.9 & 41.7 	& 45.4 & 46.9\\
    Ego-only &\XSolidBrush &	33.6 & 37.1 	& 40.3 & 35.8	\\
    Ego-only &\Checkmark  &34.6	& 38.7    & 45.6 & 41.8\\
    Ego-only& Center  & 25.4 & 22.8&		24.7 & 24.2  \\
    \midrule
    EgoExo  &\XSolidBrush & 		42.9 & 45.4 & 49.0 & 45.3\\
    EgoExo &\Checkmark &		\textbf{47.9} & \textbf{48.8} & \textbf{55.3} & \textbf{51.1}\\
    \bottomrule
  \end{tabular}
}
\vspace{-0.5pt}
\caption{Association accuracy in the cross-view association benchmark. In the \textit{fine-tuned} setting, we adopt three kinds of data sources for training, {\em i.e.}, ego-only, exo-only, and hybrid ego-exo data. By leveraging gaze information during training, the model outperforms the baseline (w/o gaze) and the center-crop counterpart.}
\vspace{-1em}
\label{tab:association}
\end{table}

%% file: tab/cross_action_under.tex
\begin{table}[t]
\centering
\scriptsize
\setlength{\tabcolsep}{5.5pt}
  \begin{tabular}[t]{lccccccc}
    \toprule
     \multirow{3}{*}{Method}&  \multirow{3}{*}{Gaze} & \multicolumn{4}{c}{Anticipation$\uparrow$} & \multicolumn{2}{c}{Planning$\downarrow$} \\
     \cmidrule(r){3-6}\cmidrule(r){7-8}
     & & Ego-V   & Ego-N & Exo-V & Exo-N & Ego & Exo  \\
    \midrule
    Exo-only & \XSolidBrush & 29.9 & 23.6 & \textbf{40.9} & \textbf{40.5} & 84.7 & \textbf{76.1}  \\
    Ego-only & \XSolidBrush & 33.4 & 37.8 & 28.9 & 17.6 & 83.4 & 84.5  \\
    Ego-only & \Checkmark & \textbf{40.5} & \textbf{52.8} & 37.6 & 37.6 & \textbf{80.0} & 82.6  \\
    Ego-only & Center & 33.2 & 38.6 & 34.1 & 32.7 & 82.6 & 84.7  \\
    \midrule
    \multicolumn{8}{l}{\emph{Unsupervised Domain Adaption}} \\
    Ego2Exo & \XSolidBrush & 33.6 & 38.1 & 35.4 & 28.7 & 83.0 & 84.1  \\
    Exo2Ego & \XSolidBrush & 30.4 & 23.6 & \textbf{39.2} & 39.8 & 83.9  & 79.0  \\
    Ego2Exo & \Checkmark & \textbf{40.8} & \textbf{54.2} & 38.7 & 37.1 & 82.8 & 84.3  \\
    Exo2Ego & \Checkmark & 33.5 & 31.3 & 39.1 & \textbf{40.1} & \textbf{82.4} & \textbf{78.8}  \\
    \midrule
    \multicolumn{8}{l}{\emph{Knowledge Distillation}} \\
    Ego2Exo & \XSolidBrush & 29.6 & 24.9 & \textbf{41.6} & \textbf{45.2} & 84.3 & 75.5 \\
    Exo2Ego & \XSolidBrush & 34.0 & 38.4 & 28.6 & 18.6& 83.1 & 84.3 \\
    Ego2Exo & \Checkmark & 29.9 & 25.0 & 41.2 & 45.1 & 84.8 & \textbf{75.1}\\
    Exo2Ego & \Checkmark& \textbf{41.0} & \textbf{56.1} & 37.7 & 39.1 & \textbf{79.5} & 82.6 \\
    \midrule
    \multicolumn{8}{l}{\emph{Co-training}} \\
    Ego \& Exo & \XSolidBrush & 33.5 & 37.4 & 39.6 & 44.3 & 83.2 & 76.0  \\
    Ego \& Exo & \Checkmark & \textbf{43.8} & \textbf{53.3} & \textbf{40.3} & \textbf{44.4} & \textbf{79.0}  & \textbf{75.6} \\
    \bottomrule
  \end{tabular}
\vspace{0.5pt}
\caption{Results of cross-view action anticipation and planning benchmarks. For anticipation, the class-mean Top-5 recall is used as the evaluation metric (higher is better). For planning, the Edit distance is used as the evaluation metric (lower is better).}
\vspace{-0.4cm}
\label{tab:sta-and-lta}
\end{table}

%% file: tab/skill_assessment.tex
\begin{table}[t]
\centering
\resizebox{\columnwidth}{!}{
\begin{tabular}{lccccc} 
\toprule
Method            & Gaze  & Egg Cracking & Peeling & Stir-fry & Cutting  \\ 
\midrule
\multicolumn{6}{l}{\emph{Ego pairs only}} \\
Who's better \cite{doughty2018s} & \XSolidBrush & 74.79        & 75.98   & 77.54    & 76.85    \\
RAAN \cite{doughty2019pros}         & \XSolidBrush & 78.45        & 78.52   & 82.53    & 79.09    \\
Who's better \cite{doughty2018s} & \Checkmark & 77.91        & 76.70   & 79.13    & 77.02    \\
RAAN \cite{doughty2019pros}         & \Checkmark & 82.08        & 79.37   & 83.92    & 79.36    \\ 
\hline
\multicolumn{6}{l}{\emph{Ego pairs + Exo}} \\
RAAN \cite{doughty2019pros} + RN         & \XSolidBrush & 77.92        & 77.09   & 81.54    & 78.26    \\
RAAN \cite{doughty2019pros} + TL         & \XSolidBrush & 78.81        & 79.46   & 82.50    & 78.73    \\
RAAN \cite{doughty2019pros} + RN         & \Checkmark & 82.14        & 79.32   & 83.51    & \textbf{79.44}    \\
RAAN \cite{doughty2019pros} + TL         & \Checkmark & \textbf{82.16}        & \textbf{79.59}   & \textbf{84.05}    & 79.29    \\
\bottomrule
\end{tabular}
}
\vspace{-0.5pt}
\caption{Ranking accuracy of cross-view referenced skill assessment. In the upper part of the table, only ego video pairs are used, while in the lower part, exo demonstrations are incorporated by ``RN'': relation network and ``TL'': triplet loss.
}
\vspace{-0.4cm}
\label{tab:skill_assess}
\end{table}

%% file: sec/6_conclusion.tex
\section{Conclusion}
The ability to bridge asynchronous procedural activities in ego- and exo-views is imperative for next-generation embodied AI in executing sophisticated tasks in the real world. As a fundamental step, our \dataset encompasses a rich collection of egocentric videos, each captured when replicating procedures of exocentric demonstration videos, but performed in different environments and at different times. 
This realistic setup, combined with our multimodal annotations, allows us to construct 4 novel benchmarks, serving as a versatile platform for investigating how cross-view asynchronous activities can be bridged. 
\dataset also enables new research directions \emph{e.g.,} how to better leverage gaze and hand-associated annotations. Results from the benchmarks show weaknesses of current models in bridging ego- and exo-view asynchronous activities, leaving significant room for future work to improve upon. 

\noindent\small{\textbf{Acknowledgement. }This work is supported by the National Key R\&D Program of China (No.2022ZD0160102) and the Industry Collaboration Projects Grant, Shanghai Committee of Science and Technology, China (No.22YF1461500).}

\noindent\small{Key contribution statement: Guo made key contributions to annotation processing and 3 action benchmarks. Jilan made key contributions to the association and caption benchmarks. Mingfang contributed primarily to the skill benchmark.}

%% file: sec/X_suppl.tex
\clearpage
\setcounter{page}{1}
\maketitlesupplementary
This supplementary material shows details about our benchmark including formal definition, implementation, and additional experiment results. Also, we show additional details about the collection and annotation of the dataset.
\renewcommand{\thesection}{S\arabic{section}}
\renewcommand{\thetable}{S\arabic{table}}  
\renewcommand{\thefigure}{S\arabic{figure}}
\section{Additional Benchmark Details}
\label{sec:bench}
\subsection{Cross-view association}

\subsubsection{Detailed task definition}
The training set consists of separate egocentric videos $V^{\text{ego}}$ with associated narration $T^{\text{ego}}$ and exocentric videos and narrations ($V^{\text{exo}}$,$T^{\text{exo}}$). For each egocentric video, a sequence $g$ with corresponding gaze is provided. Note that, we do not provide explicit pair information in the training set.

In the validation/test set, we introduce two evaluation settings, \emph{i.e.,} Ego2Exo and Exo2Ego. We describe the formulation of Ego2Exo as follows. Each sample consists of an egocentric query video $V^{\text{ego}}$ and $K$ exocentric candidate videos $\{V^{\text{exo}}_1,...,V^{\text{exo}}_K\}$, where only one candidate exocentric video corresponds to the query egocentric video, \emph{i.e.,} the same action is being performed. In the Exo2Ego setting, the query is exocentric videos while egocentric videos form the candidate set. 
For both Ego2Exo and Exo2Ego settings, we consider $K=20$ candidates. 

\subsubsection{Implementation details}
\noindent\textbf{Training setting.}
As explicit pairing is not available in the training set, we propose a simple baseline approach to align egocentric videos and exocentric videos in the semantic space. In specific, we train a dual-encoder architecture consisting of a video encoder $f_v(\cdot)$ and a text encoder $f_t(\cdot)$ on both ego- and exo-videos and narrations using the contrastive loss, named as \textit{co-training} in our experiments. 
Following~\cite{egovlp,lavila}, we adopt a TimeSformer-B~\cite{timesformer} as the video encoder and a clip~\cite{clip} text encoder. We randomly sample 4 frames as input. The model is initialized with weights pre-trained on Ego4d video and text pairs~\cite{ego4d,egovlp}. We train the dual encoder model for 5 epochs with a fixed learning rate 1e-5 and a batch size of 32. 
At the inference stage, the text encoder is discarded and only the video encoder is used. For each query, we compute its video representation with $K$ features of the candidate videos and select the one with highest cosine similarity as the model prediction. 

\begin{figure}
\includegraphics[width=\columnwidth]{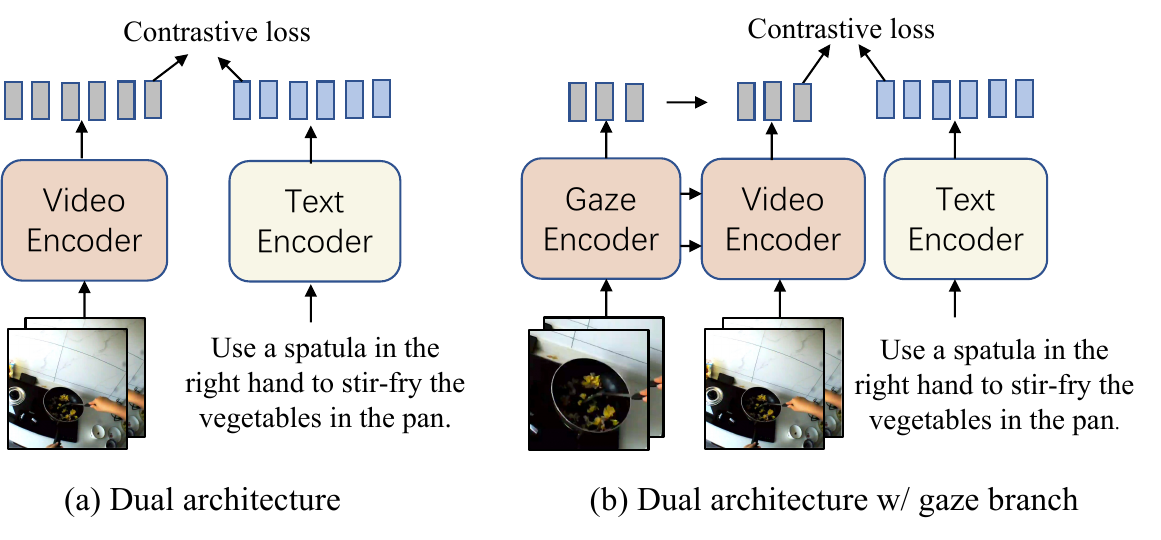}
    \caption{Cross-view association network with naive dual architecture (a) and improved architecture with additional gaze branch (b).}
    \label{fig:association}
\end{figure}

\noindent\textbf{Network architecture.}
To leverage the gaze information in associating egocentric and exocentric videos, we further propose a multi-view branch for the video encoder~\cite{multiview}. One branch encodes the original video while the other branch encodes the gaze cropped video, as illustrated in Fig.~\ref{fig:association}. The feature of the original video cross-attends to the gazed video feature every at the 5$^{\text{th}}$, 8$^{\text{th}}$, and 11$^{\text{th}}$ transformer block, enabling multi-scale feature fusion for improved visual representation. For exocentric videos, we simply input the original video to the gaze branch.

\subsubsection{Annotation details}
The process of pair construction consists of five stages:
(1) Scenario Matching. 
We gather all the egocentric and exocentric videos under the same scenario (e.g. cooking the same dish or conducting the same experiment) into each group. 
(2) Noun and Verb Matching. 
Based on the noun and verb vocabularies, for each group, we pair an egocentric caption with another if they contain exactly the same nouns and verbs. 
(3) Sentence Matching with LLM. 
We ask the LLM (e.g. ChatGPT) to determine whether each ego-exo caption pair obtained in stage 2 describes the same activity at sentence-level, reducing the linguistic ambiguity caused by word matching. 
(4) Negative Sampling. 
We randomly choose video clips from the same video as negative samples in the candidate set.
(5) Two-round Manual Verification. 
We manually check the semantic meaning of each ego-exo pair and corresponding ego-exo video to make sure the exact match. This verification is performed in two rounds by two different individuals. In total, the size of the validation/test set is 868/2200. As stated in the main manuscript, we do not provide such pairs for the training set and leave the modeling of cross-view association on unpaired samples to be further explored for the community.



\subsection{Cross-view action anticipation \& planning}

\subsubsection{Detailed task definition}
\label{sec:cv-anti-plan-definition}

Task definitions of cross-view action anticipation and planning have followed the previous benchmarks of \cite{epickitchen} and \cite{ego4d}. Our cross-view benchmark extends on the original task setting and focuses on mutual assistance between egocentric and exocentric video data. 

\minisection{Action anticipation.} 
The action anticipation task focuses on forecasting the verb and noun categories of the subsequent fine-level action at $\tau=1$ second into the future.
Considering a fine-level action segment $a=(s,e,c)$, where $s$, $e$, and $c$ represent the start time, end time, and category of $a$ respectively, the model is restricted to observing video data only up to time $s-\tau$. 
The model's objective is to predict the forthcoming action, encompassing relevant verbs and nouns.
The performance of the model in this benchmark is evaluated using class-mean Top-5 recall, as outlined in~\cite{epickitchen}.

\minisection{Action planning.}
The objective of the action planning task is to generate the next $K$ steps of coarse-level actions. Considering $N_a$ fine-level action segments $A=\{a_i=(s_i,c_i)\}_{i=1}^{N_a}$, where $s_i$ (ensuring $s_i<s_{i+1}$) and $c_i$ represent the start time and category of $a_i$ respectively, the model is limited to observing video data up to time $s_i$ and is tasked with forecasting the $K$ actions $s_i,..., s_{i+K-1}$ into the future.
For evaluation purposes, we adopt ED@$K$ as the metric, following the approach outlined in Ego4D LTA~\cite{ego4d}. 
In our specific configuration, we set $K$ to 8 and sample 5 predicted sequences for evaluation.

\begin{figure}
\includegraphics[width=\linewidth]{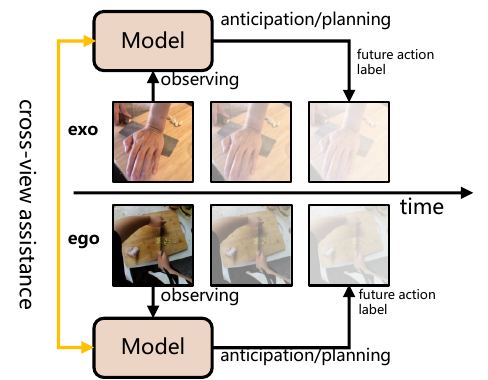}
    \caption{Overall framework of cross-view action anticipation and planning. The model observes the past video and tries to anticipate the next fine-level action (action anticipation) or the next $K$ steps of the coarse-level actions (action planning). The model gets assistance from the knowledge in the other view.}
    \label{fig:anti-plan-cv-main}
\end{figure}

\minisection{Cross-view benchmark.} In our cross-view benchmark, we begin by assessing zero-shot cross-view action understanding. Following this, we employ various methods to leverage information in one view to assist the understanding in the other view. Thus, this benchmark is focused on designing approaches that utilize both ego and exo-view data to enhance the cross-view performance. Figure~\ref{fig:anti-plan-cv-main} shows the overall framework of our cross-view benchmark for action anticipation and planning. Figure~\ref{fig:anti-plan-cv} further illustrates our various cross-view settings.

\subsubsection{Implementation details}
\label{sec:anti-plan-impl}

\begin{figure*}[t]
\includegraphics[width=\linewidth]{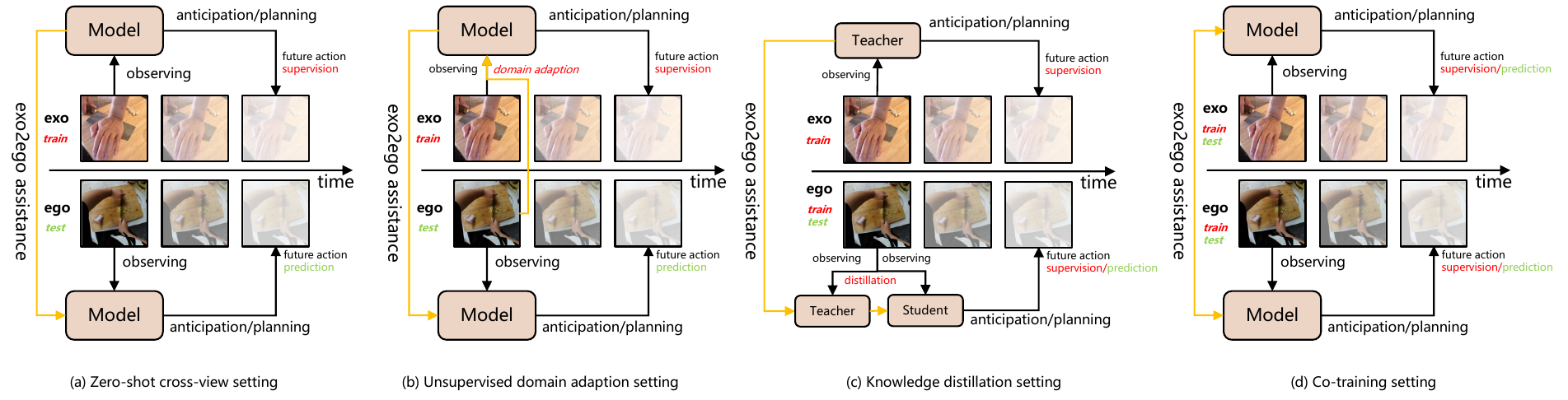}
    \caption{Four settings for cross-view action anticipation and cross-view action planning. (a) The zero-shot setting directly evaluates the model trained on one view on the test data of the other view. (b) The unsupervised domain adaptation (UDA) setting involves leveraging data from another view, but without using the labels associated with this data. (c) In the knowledge distillation setting, for a model in one view, a teacher model trained on the other view is used to provide assistance. (d) The co-training setting directly uses the data and labels of both views. (a) to (d) represent 4 increasing degrees of cross-view information usage.}
    \label{fig:anti-plan-cv}
\end{figure*}

\minisection{Network architecture.} 
To adapt our cross-view training settings, we rely on the TA3N~\cite{chen2019ta3n} code base, acknowledged for its clarity and comprehensibility, and widely adopted in recent research.
We employ CLIP~\cite{clip} as the feature extractor for generating frame-level video features.
Both action anticipation and planning tasks entail leveraging historical information to forecast future actions. Thus, we input a 2-second context into the model. Within the specified temporal range, we uniformly sample 5 frames as the input. Utilizing the 3D feature map extracted by TA3N~\cite{chen2019ta3n}, we perform average pooling to condense the feature map into a vector $v \in \mathbb{R}^{d}$. We employ a projector $\mathbf{W}_{anti}$ to predict $C_{anti}$ classes for the action anticipation task, where $C_{anti}$ is the number of verb or noun categories. For the action planning task, we use a projector $\mathbf{W}_{plan}$ to predict $C_{plan} \times K$ classes, where $C_{plan}$ is the number of coarse-level categories and $K$ (set to $8$) is defined in Sec~\ref{sec:cv-anti-plan-definition}.

\minisection{Training.} We first introduce the training settings of both tasks. Given the anticipation logits $y_{anti}$ produced by the model and the corresponding ground truth $\hat{y}_{anti}$, we employ the standard cross-entropy loss for supervision:

\begin{equation}
\mathcal{L}_{anti} = \mathcal{L}_{CE}(y_{anti}, \hat{y}_{anti}).
\end{equation}

For an action sequence $y_{plan}^{1}, \ldots, y_{plan}^{K}$ predicted by the action planning model, the loss function is defined as:

\begin{equation}
\mathcal{L}_{plan} = \frac{1}{K}\sum_{i=1}^{K}\mathcal{L}_{CE}(y_{plan}^{i}, \hat{y}_{plan}^{i}).
\end{equation}

The model is trained using the SGD optimizer with a learning rate set to 1e-2 and the training process spans 40 epochs.

\minisection{Zero-shot cross-view setting.} In the zero-shot cross-view setting, the model is initially trained on data in one view and directly tested on data in the other view. This setting is crucial for understanding how well a model trained on data from one perspective can adapt to and accurately interpret data from another perspective, without any additional training specific to that new viewpoint. Figure~\ref{fig:anti-plan-cv}(a) illustrates the procedure of the ``exo2ego" cross-view setting, where the model is first trained on exocentric data and then tested on egocentric data. The ``ego2exo'' setting works vice versa.

\minisection{Unsupervised domain adaptation setting.} In the unsupervised domain adaptation setting, the training process involves using data and labels from the source view, plus the video data from the target view. The annotations from the target view are not used. Figure~\ref{fig:anti-plan-cv}(b) illustrates the ``exo2ego" cross-view setting, where exocentric data serves as the source domain, and egocentric data serves as the target domain. In addition to task supervision, the overall loss function also contains a domain adaption loss derived from TA3N~\cite{chen2019ta3n} for unsupervised domain adaptation settings:
\begin{equation}
\begin{aligned}
&\mathcal{L}_{DA} = \frac{1}{N_S}\sum_{i=1}^{N_S}\mathcal{L}_{y}^i+\frac{1}{N_{S \cup  T}}\sum_{i=1}^{N_{S \cup T}}\gamma\mathcal{L}_{ae}^i \\
&-\frac{1}{N_{S \cup T}}\sum_{i=1}^{N_{S \cup T}}(\gamma^s \mathcal{L}_{sd}^i+\gamma^r \mathcal{L}_{rd}^i+\gamma^t \mathcal{L}_{td}^i).
\end{aligned}
\end{equation}

Therefore, the overall loss function for both tasks under this setting is 
\begin{equation}
\mathcal{L}=\mathcal{L}_{anti/plan}+\mathcal{L}_{DA}.
\end{equation}

\minisection{Knowledge distillation setting.}
In the knowledge distillation setting, the training process comprises two stages: (1) training the teacher model on the data in one view, and (2) training the student model on the data in the other view, meanwhile distilling knowledge from the teacher model.
Figure~\ref{fig:anti-plan-cv}(c) depicts the ``exo2ego'' cross-view setting, where the teacher model is trained on exocentric data, and the student model is trained on egocentric data. 
In addition to task supervision, the overall loss function for training the student model also contains a knowledge distillation loss for knowledge distillation settings. Specifically, we use L2 loss to minimize the feature $y_{feat}^{S}$ and $y_{feat}^{T}$ output by the student and teacher:

\begin{equation}
\mathcal{L}_{KD}=\mathcal{L}_{L2}(y_{feat}^{S},y_{feat}^{T}).
\end{equation}

Therefore, the overall loss function of the teacher and student for both tasks under this setting is 

\begin{align}
\mathcal{L}_{teacher}=&\mathcal{L}_{anti/plan}, \\
\mathcal{L}_{student}=&\mathcal{L}_{anti/plan}+\mathcal{L}_{KD}.
\end{align}

\minisection{Co-training setting.} In the co-training setting, exocentric and egocentric data are both used to train the model. The model is then evaluated on the test set of the egocentric and exocentric data.
Figure~\ref{fig:anti-plan-cv}(d) depicts the ``exo \& ego'' co-training setting.

\subsubsection{Annotation details}
\label{sec:fine-action-annotation}

\minisection{Cross-view action anticipation.} The annotation process for cross-view action anticipation involves three stages: (1) extracting verbs and nouns for each fine-level action clip, (2) aligning the closed categories of training, validation, and testing set across egocentric and exocentric videos, (3) restricting the closed set to the intersection of categories present in all egocentric and exocentric videos, and (4) managing the long-tail distribution of the data by filtering out categories that occur less than \nicefrac{1}{100} of the highest occurrence category. We delete all video clips without any label. As a result, this task contains 19/31 verb/noun categories. The size of the egocentric train/validation/test set is 34.5k/7.7k/17.3k, and the size of the exocentric train/validation/test set is 6.1k/2.1k/4.8k. 

\input{tab/cross_action_under_val}

\minisection{Cross-view action planning.} Cross-view action planning utilizes coarse-level annotations with a total of 27 classes for training, validation, and testing. We sort all action steps in each video by their start time. Consequently, this task is oriented towards predicting potential sequences of future action starts. 
After filtering, we obtain 2.1k/0.8k/1.2k action steps in the egocentric train/validation/test set and 2.4k/0.3k/0.4k action steps in the egocentric train/validation/test set. Note that it is also possible to use the fine-level action annotations for this task, which will result in a much larger dataset split. We do not use this setting since we observe a large variation in the fine-level actions due to practical issues such as environmental constraints and unskilled performance. We believe the combination of our cross-view anticipation and cross-view planning can well evaluate the ability to bridge ego-exo procedural activities at both clip-level and task-level.

\subsubsection{Additional results}

Table~\ref{tab:sta-and-lta-val} presents the results of our baseline models on the validation set for the cross-view action anticipation and planning benchmarks. 
In the first block, zero-shot cross-view evaluation (\textit{e.g., Exo-only} evaluated on ego view, and \textit{Ego-only} evaluated on exo view)  results in the lowest performance levels. This outcome underscores the challenge of applying learned representations from one perspective directly to another without any intermediary processing or adaptation. A significant improvement in zero-shot cross-view performance is observed with the introduction of gaze-cropped inputs. This enhancement suggests that \textbf{gaze can be an effective bridge for the ego and exo actions}. Further improvements in performance are noted when implementing methods such as Unsupervised Domain Adaptation (UDA), Knowledge Distillation (KD), and Co-Training (CT). The results also demonstrate that the extent of performance improvement varies across different cross-view settings. This variation highlights the complexity of bridging activities in ego and exo views and the importance of selecting the most appropriate method based on the specific requirements of each task.

\begin{figure}
\includegraphics[width=\linewidth]{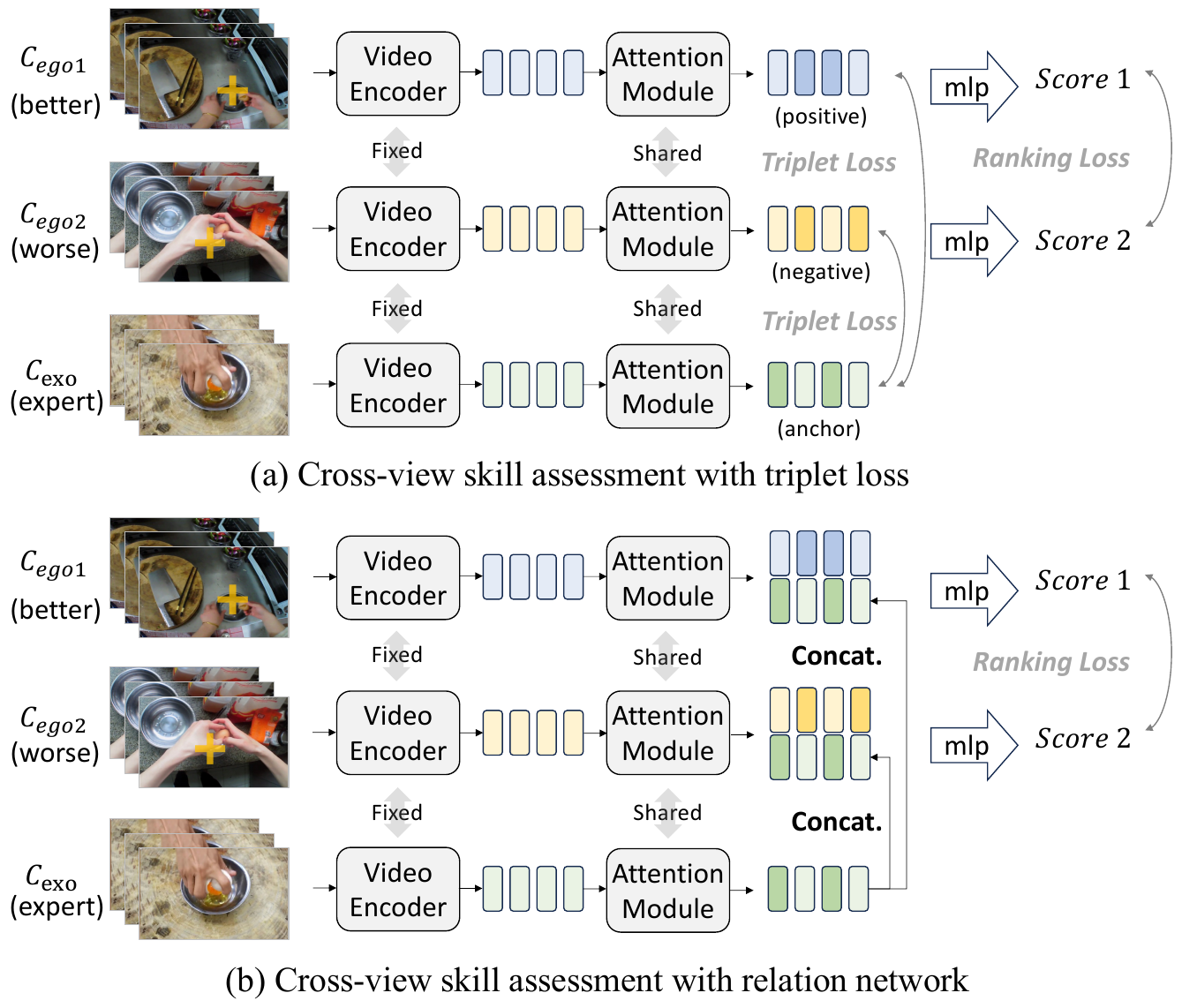}
    \caption{Cross-view referenced skill assessment with triplet loss and relation network.}
    \label{fig:skill_pipe}
\end{figure}

\subsection{Cross-view referenced skill assessment}

\subsubsection{Detailed task definition}
Our training dataset comprises the following components: (1) Egocentric Video Pairs: Denoted as $P$, each pair $(C_{ego1}, C_{ego2}) \in P$ is arranged such that video $C_{ego1}$ displays better skill than $C_{ego2}$. (2) Accompanying Gaze Sequences: For every pair of egocentric videos $(C_{ego1}, C_{ego2}) \in P$, corresponding gaze sequences $(g_1, g_2)$ are provided. (3) Exo-View Expert Demonstration: Each pair $(C_{ego1}, C_{ego2}) \in P$ is accompanied by an expert demonstration video $C_{exo}$, showcasing the same action as in $C_{ego1}$ and $C_{ego2}$ from an exo-view perspective. The objective is to develop a ranking function $f(\cdot)$ that adheres to the condition $f(C_{ego1}) > f(C_{ego2})$ given $(g_1,g_2)$ and $C_{exo}$ as the reference.

\subsubsection{Implementation details}

\minisection{Network architecture.} 
As shown in \cref{fig:skill_pipe}, we assume $C_{ego1}$ exhibits a higher skill level compared to $C_{ego2}$. Built upon a pairwise ranking skill assessment model RAAN~\cite{doughty2019pros}, we employ different video encoders, including I3D~\cite{carreira2017quo} and VideoMAE~\cite{tong2022videomae}, to extract video features from $C_{ego1}$, $C_{ego2}$, and $C_{exo}$. The features are processed by an attention module as described in~\cite{doughty2019pros} and resulting in refined features $F_{ego1}$, $F_{ego2}$, and $F_{exo}$. Then, we apply two different approaches to leverage the reference exo-view demonstration video: 
1) Triplet loss (TL). We designate $F_{exo}$ as the \textit{anchor}, $F_{ego1}$ as the \textit{similar item (positive)}, and $F_{ego2}$ as the \textit{dissimilar item (negative)}. Then, we apply a triplet margin loss with margin = 1, to aid the model in understanding that the anchor is closer to the positive than the negative item. In our scenario, $C_{ego1}$ demonstrates a skill level closer to the expert.
2) Relation network (RN). 
Inspired by \cite{sung2018learning}, we implement a relation network that concatenates the features of the ego and exo clips. Precisely, we set $F_{ego1} = Concat(F_{ego1}, F_{exo})$ and $F_{ego2} = Concat(F_{ego2}, F_{exo})$. By combining ego and exo features, this network is designed to implicitly discern which of the two egocentric video clips bears a closer relation to the demonstration video in terms of skill level. 
Finally, the refined features $F_{ego1}$ and $F_{ego2}$ are processed by an MLP to regress skill scores for the two ego videos.

\minisection{Training.} 
For the ego branch of our network, we employ the training objectives from \cite{doughty2018s, doughty2019pros}. 1) a margin ranking loss is applied on the finally generated scores to ensure $ego1$ is ranked higher than $ego2$. 2) a disparity loss is applied within the attention module to prevent the network from getting trapped in local minima during training 3) a rank-aware loss and a diversity loss are also applied following \cite{doughty2019pros}. Besides the ego branch, to leverage the exo demonstration video, we propose to utilize a triplet loss to aid the model in comprehending that $ego1$ exhibits skills more akin to those of an expert.

\input{tab/skill_dataset_compare}

\input{tab/skill_assessment_supp}

\subsubsection{Annotation details}
We include two types of annotations for skill level. The first type is self skill assessment.
During data collection, subjects are asked to assess themselves on various aspects, including their familiarity with cooking environments, the number of times they have completed the task previously, the frequency of performing the task, the typical duration required to complete the task, and whether they've taught others how to perform the task. Based on the self-evaluation results, we have observed a considerable diversity in subjects' skill levels, which motivates us to craft the skill assessment benchmark. One related work is HoloAssist~\cite{HoloAssist2023} where they show the distribution of the performers’ familiarity with the tasks measured by a self-reported score (0-10) by the subjects. However, no related benchmarks is provided by HoloAssist. 

\input{tab/cross_view_tas}

One drawback of self-evaluated skill level is that individuals may showcase varying skill levels in each video instance, even across multiple attempts~\cite{doughty2018s}. As a more objective complement of the self-assessment,  we adopt the pairwise comparison approach \cite{doughty2018s,doughty2019pros,li2019manipulation} for annotation. We provide annotators with four criteria: Fluency, Speed, Proficiency, and Skillfulness. These standards serve as the basis for their ranking assessment. From the annotation results, we find 40\% of the rankings deviate from the rankings based on the self-evaluations of the two subjects in the video pair. This finding supports that relying solely on self-evaluation is inadequate for creating a robust skill assessment benchmark.

As shown in Table~\ref{tab:skill_datasets}, our dataset stands out as the only skill assessment dataset featuring the gaze modality and corresponding exo-view demonstration videos. Notably, our dataset surpasses previous ones in both video clip quantity and valid pair numbers. We follow the setting in~\cite{doughty2018s, doughty2019pros} to employ 4 individuals to rank the same video pair to ensure credibility. We exclude annotations with fewer than 3 consistent opinions instead of 4 to ensure our dataset contains challenging pairs. Regarding action categories, our dataset comprises 6 actions: Egg cracking, Peeling, Stir-fry, Cutting into chunks, Slicing into strips, and Chopping into pieces. In the main paper, we merge the last three actions into a comprehensive category labeled ``Cutting'', encompassing various knife-using skills.

\subsubsection{Additional results}
Results with I3D~\cite{carreira2017quo} feature are shown in Table 5 of the main manuscript. We show the results with VideoMAE~\cite{tong2022videomae} feature in \cref{tab:skill_assess_supp}. Comparing results from the two tables, we observe an overall increase in performance in all cases in Table~\ref{tab:skill_assess_supp} because of the stronger backbone model. While we can still observe performance gain when adding Exo reference video, this improvement is less significant compared with the corresponding table in the main manuscript. We suspect that this variation is attributed to the varying degrees of influence that the intrinsic properties of the extracted features exert on the observed enhancements.

\subsection{Cross-view action segmentation}

\subsubsection{Detailed task definition}
The action segmentation task in our framework is focused on both categorizing each time step and delineating action steps within procedural videos. Given a lengthy video $V$ comprising $N_V$ frames at 25 FPS, the model is tasked with classifying the category of each frame in the video. The evaluation metric includes assessing frame-level classification accuracy. Additionally, sequence-level metrics such as edit distance and instance-level metric F1 are employed for further evaluation~\cite{farha2019ms}. The extended cross-view action segmentation benchmark, similar to cross-view action anticipation and planning, aims to pursue performance improvement by receiving aid from other views.

\subsubsection{Implementation details}

\minisection{Network structure.} 
We employ I3D~\cite{carreira2017quo} as the feature extractor to generate temporal features, following the methodology of previous work~\cite{farha2019ms}. To implement our various training settings, we utilize the SSTDA~\cite{chen2020sstda} code base. For both training and testing, we downsample feature sequences and label sequences by a factor of 5 for efficiency.

\minisection{Training.} 
The loss function used to train the action segmentation task is derived from SSTDA~\cite{chen2020sstda}. The model consists of multiple stages. The overall loss function for a single stage is a combination of the classification loss and smoothing loss:
\begin{align}
\mathcal{L}_{seg}=&\mathcal{L}_{cls} + \gamma \mathcal{L}_{smooth}.
\end{align}

The model is trained using the Adam~\cite{kingma2014adam} optimizer with a learning rate set to 1e-3 and the training process spans 150 epochs.

\minisection{Cross-view settings.}
Similar to cross-view action anticipation and planning, action segmentation also performs four cross-view settings. Though cross-view action segmentation shows different input and output, which yields dense prediction, the implementation of cross-view settings is consistent with Section~\ref{sec:anti-plan-impl}.

\subsubsection{Annotation details}

The annotation for the cross-view action segmentation task is derived from coarse-level annotations. To create non-overlapping segment annotations for temporal action segmentation, we establish the center point of the overlapping portion of two segments as their boundary. Subsequently, we introduce background segments labeled as ``no action" in temporal regions not covered by action annotations. Finally, we obtain 173/57/85 videos in the egocentric train/validation/test set and 210/24/32 videos in the exocentric train/validation/test set.

\subsubsection{Experimental results}

Table~\ref{tab:as} presents the results of our baseline models on the validation set and test set for the cross-view action segmentation benchmarks. These results mirror the trends observed in the action anticipation and planning benchmarks: without any assistance from another view, the models can only perform well on the test data in the same view. The inclusion of gaze data enhances model performance in both the ego-only setting and the cross-view setting. This suggests that focusing on areas of visual attention, as indicated by gaze data, is beneficial for better understanding and segmenting actions, regardless of the viewpoint. When information from another view is leveraged, all three methods – Unsupervised Domain Adaptation (UDA), Knowledge Distillation (KD), and Co-Training (CT) – contribute to performance improvements in the cross-view setting. Each method offers a different mechanism for integrating cross-view insights, thus aiding in the segmentation task. Reflecting the varying degrees of labeled data utilization, Co-Training (CT) tends to outperform Knowledge Distillation (KD), which in turn outperforms Unsupervised Domain Adaptation (UDA).

\subsection{Cross-view referenced video captioning}
\begin{figure}
\includegraphics[width=\columnwidth]{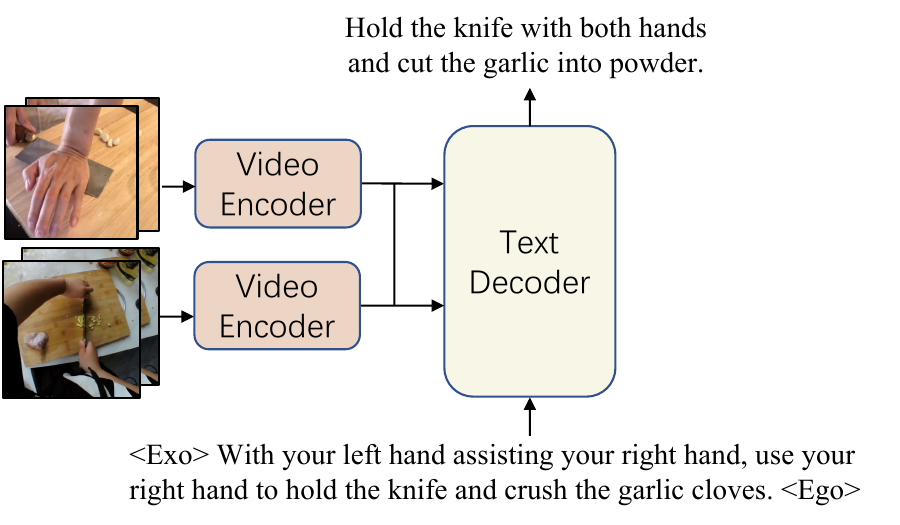}
    \caption{Cross-view referenced captioning with a video encoder and a text decoder.}
    \label{fig:captioning}
\end{figure}
\subsubsection{Detailed task definition}
Cross-view referenced video captioning evaluates the model's captioning ability to leverage cross-view information for caption generation. 
Our motivation is that egocentric videos require extensive efforts to collect, and are thus limited in scale and diversity. In contrast, large-scale exocentric videos can be easily sourced from the Internet. The question is, \emph{how to leverage such exocentric videos to help the understanding of limited egocentric videos?}. 

Formally, at the training stage, we have egocentric videos of limited size $\{(V_1^{\text{ego}},T_1^{\text{ego}}),...,(V_N^{\text{ego}},T_N^{\text{ego}})\}$ with $N$ samples, and exocentric videos $\{(V_1^{\text{exo}},T_1^{\text{exo}}),...,(V_M^{\text{exo}},T_M^{\text{exo}})\}$, where $N\ll M$. Each video is paired with a fine-grained text description. 
The goal is to train a cross-view video captioning model $f(\cdot)$ using exocentric videos as references. At the inference stage, the model is required to generate the captions of the testing egocentric videos, given the other set of exocentric videos as references. Note that, $N\leq M$ only holds for the training set. 
In particular, we limit the number of the referenced exocentric videos by formulating the task as a $K$-shot captioning~\cite{flamingo} problem, where $K$ denotes the maximum number of exocentric videos that the model is allowed to use during inference. 
The inference process can be formulated as $f(V^{\text{ego}}|\{(V_1^{\text{exo}},T_1^{\text{exo}}),...,(V_K^{\text{exo}},T_K^{\text{exo}})\}$
In practice, we consider three settings, 0-shot, 1-shot, and 2-shot. 

\subsubsection{Annotation}
We directly apply the fine-grained language annotations in our dataset. The referenced exocentric videos are randomly selected for training/validation/testing, respectively. 
The training set only contains 1000 egocentric videos with 6270 referenced exocentric videos. 
For the validation/testing set, there are 8181/2143, 18243/4930 
egocentric videos and referenced exocentric videos, respectively.
\input{tab/cross_view_captioning}
\subsubsection{Implementation details}
For the baseline model, we choose a Flamingo-style captioning model~\cite{flamingo,openflamingo,otter}, an advanced vision-language model designed for few-shot vision-language tasks, as shown in Fig.~\ref{fig:captioning}. Please refer to~\cite{flamingo} for the architectural details. We simply pre-pend the referenced video(s) before the input video, and add the referenced caption as prompts to the text decoder. We train the model for 3 epochs using the Adam optimizer, with an initial learning rate of 1e-4 and a batch size of 32. We adopt the cross-view association network (Fig~\ref{fig:association}(b)) to select referenced samples.

\subsubsection{Results}

Table~\ref{tab:captioning} lists the cross-view referenced captioning performance. We consider three baseline models: (i) \textbf{Single-view} models include \textit{Ego-only} and \textit{Exo-only}, where the former one merely adopts egocentric videos for training and inference without seeing exocentric videos. The \textit{Exo-only} model uses all referenced exocentric videos for training, and it is then evaluated on egocentric videos. (ii) \textbf{Co-training model} is trained on both egocentric videos and referenced exocentric videos, and transferred to egocentric test videos. 
(iii) \textbf{Referenced-training} model refers to our model introduced in Fig.~\ref{fig:captioning}, where the model leverages one (1-shot) or two (2-shot) exocentric videos to make predictions. 
As shown in Table~\ref{tab:captioning}, both the co-training model and referenced-training models outperform single-view models. For co-training models, the performance gain is due to the increased number of training data (ego+exo), compared to ego-only and exo-only counterparts. In terms of referenced-training models, they generally outperform the ego-only counterpart by additionally incorporating exocentric videos in the model. Results of both the co-training model and referenced-training models indicate the effectiveness of utilizing exocentric videos in improving egocentric video captioning when the data is of limited scale. 

\subsection{Zero-shot action recognition}


We assess the zero-shot classification performance of verb and noun subsets. In cases where samples have multiple labels, we straightforwardly replicate the samples for testing. Our testing procedure follows CLIP~\cite{clip}, evaluating the vision-language models based on Top-1 and Top-5 accuracy. 

\subsubsection{Annotation}
In this task, our evaluation specifically addresses zero-shot transfer within the closed set and does not encompass cross-view settings. 
It is noteworthy that this annotation does not require ensuring consistent categories between egocentric and exocentric datasets across their respective validation and testing sets. The size of the resulting egocentric verb-validation/verb-test/noun-validation/noun-test set is 14.4k/32.6k/20.2k/44.8k, and the size of the exocentric verb-validation/verb-test/noun-validation/noun-test set is 4.2k/10.4k/5.7k/13.1k, respectively.

\subsubsection{Implementation details}
We use the 16 prompts from the zero-shot classification on Kinetics~\cite{carreira2017quo} for verb and noun subsets. 
These prompts are listed in Table~\ref{tab:AR-ZS-prompts}. 
We sample the center frame of each video clip, and use OpenAI CLIP~\cite{clip} to extract the visual features and textual features.


\subsubsection{Experimental results}

Table~\ref{tab:AR-ZS} shows the performance of zero-shot action recognition. \textit{Oracle} is the upper bound of accuracy, given that this is a multi-class action recognition problem. On both the validation set and the test set, the zero-shot performance on egocentric videos is worse than that on exocentric videos, particularly in the top-1 accuracy. This result indicates the limitation in cross-view action understanding of the current method. 

\input{tab/action_recognition_zs}

\input{tab/action_recognition_zs_prompts}

\subsection{Fine-tuned action recognition}

\subsubsection{Detailed task definition}

We formulate the conventional Fully-supervised setting to a multi-label classification task. In assessing the performance of fully supervised action recognition, we employ the class-wise multi-label mean Average Precision (Marco mAP) evaluation metric due to the presence of multiple labels per clip. This evaluation protocol is reasonable because it matches the long-tail attribution of actions in \dataset.

\input{tab/action_recognition_ft}

\subsubsection{Annotation}

In this task, our evaluation focuses on the closed set and does not consider cross-view settings. 
Thus, our annotations ensure that egocentric and exocentric datasets maintain consistent categories across their respective training, validation, and testing sets. 
At last, this task contains 81/211 verb/noun categories in the egocentric set and 69/183 verb/noun categories in the exocentric set.  
The size of the egocentric train/validation/test set is 36k/8k/18k, and the size of the exocentric train/validation/test set is 6.2k/2.1k/4.8k. 

\subsubsection{Implementation details}
For evaluating this task, we utilize SlowFast-R50~\cite{feichtenhofer2019slowfast} and MViT-Small~\cite{fan2021mvit} as the backbones. The weights pretrained on the Kinetics~\cite{carreira2017quo} dataset are employed for both backbones. Frames within each action clip are uniformly sampled and fed into the backbone. The multi-label classification task is supervised using the standard cross-entropy loss. Table~\ref{tab:ar-ft-param} lists the training hyperparameters.

\input{tab/action_recognition_ft_param}

\begin{figure*}
    \centering
    \includegraphics[width=\linewidth]{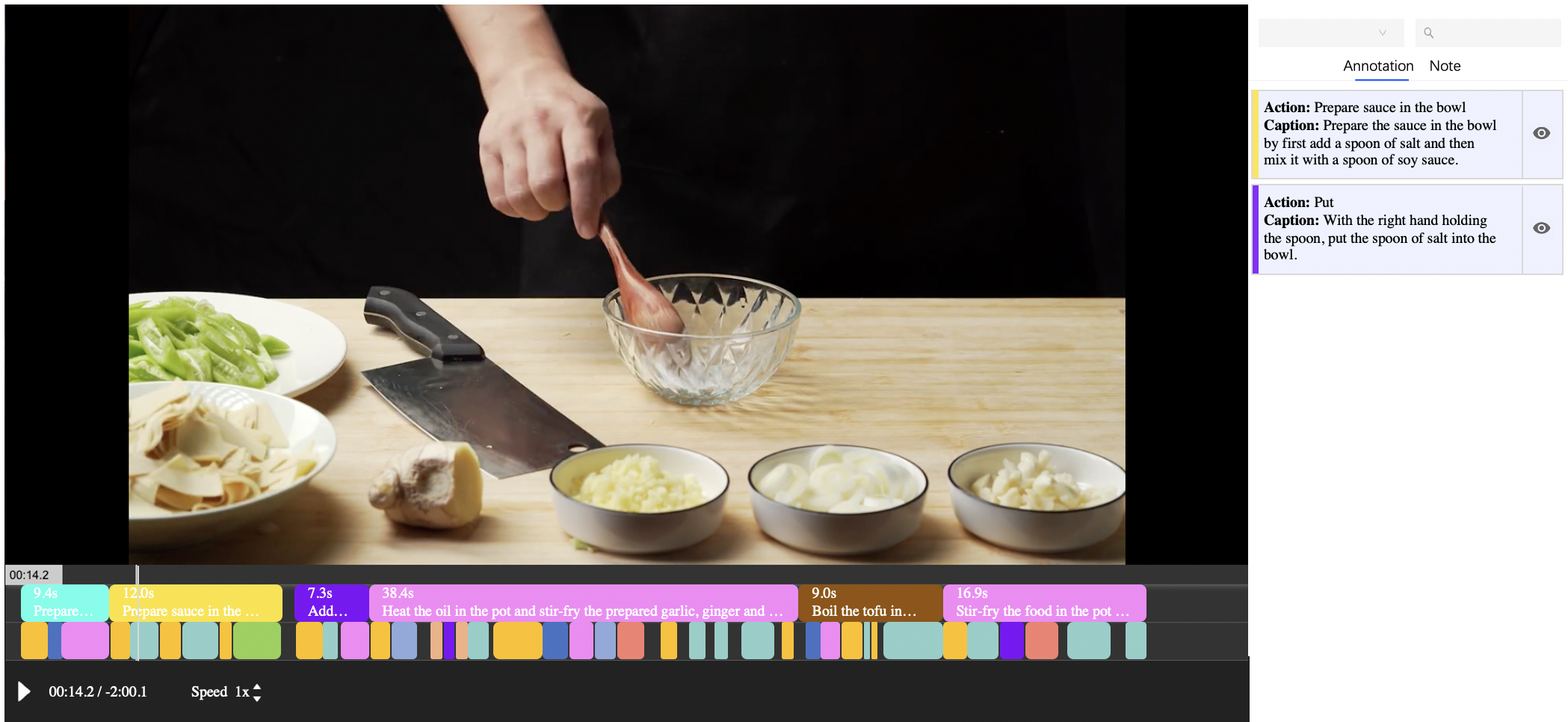}
    \caption{We use a web-based language annotation interface for the annotators. Annotators mark a segment of the video, select a category for this segment, and describe the segment based on the annotation requirement in their mother language.}
    \label{fig:interface}
\end{figure*}

\subsubsection{Experimental settings and results.}
Table~\ref{tab:AR-finetune} shows the result of fine-tuned action recognition. MViT-S~\cite{fan2021mvit} (with 16 frames input) exhibits superior performance and generalization compared to the R50-based SlowFast~\cite{feichtenhofer2019slowfast} (with 4 frames for the slow branch and 32 frames for the fast branch as input). The results in Table~\ref{tab:AR-finetune} also reveal great potential improvement on more sophisticated model structures for this dataset.

\section{Additional Dataset Details}
\label{sec:data}

\subsection{Language annotation}

Different from previous datasets~\cite{ego4d,HoloAssist2023,epickitchen}, our dataset includes two-level language annotations with manually annotated temporal boundaries.
As described in Section 3.2 of the main manuscript, our annotation includes a coarse-level language annotation and a fine-level language annotation. We designed a web-based interface to facilitate the annotation. An example screenshot is shown in Figure~\ref{fig:interface}.

\begin{figure*}
    \centering
    \includegraphics[width=\linewidth]{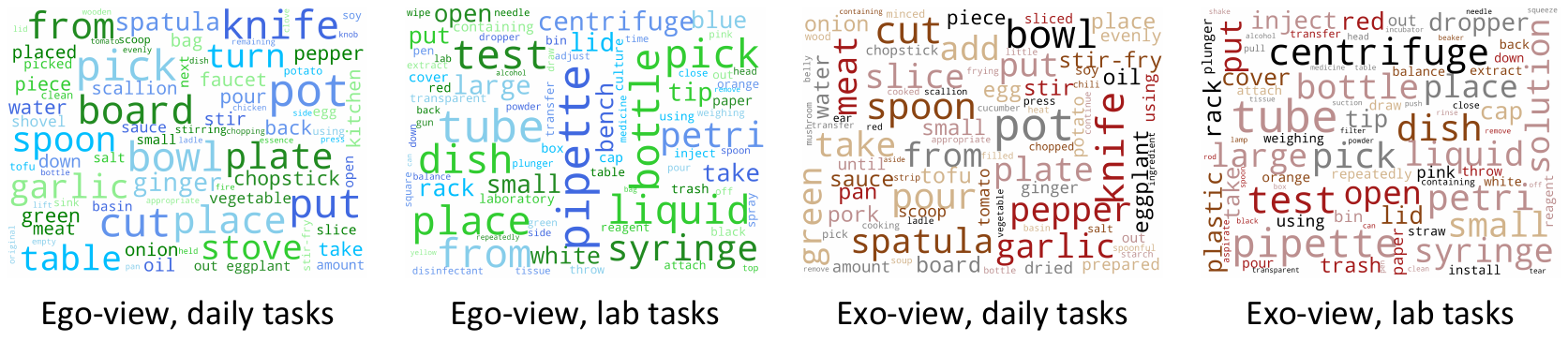}
    \caption{Word cloud of annotations separated by views and tasks.}
    \label{fig:cloud}
\end{figure*}

\begin{figure}
    \centering
    \includegraphics[width=\linewidth]{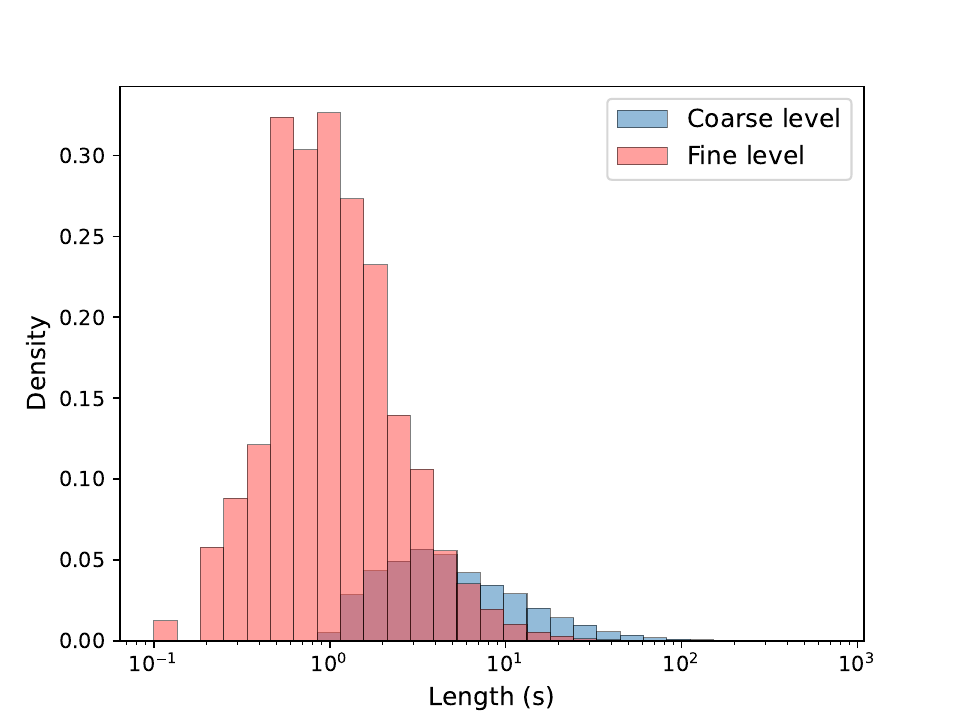}
    \caption{Distribution of the lengths of the coarse-level and fine-level language annotations. }
    \label{fig:length_dist}
\end{figure}

For each video, the annotators are asked to quickly skim the video to grab the overall content, and then begin the annotation of each session. For the daily tasks, the annotators are instructed to describe each segment based on their own knowledge. For the tasks in specialized laboratories, we train the annotators showing them the process of the experiments, the technical terms of some tools/reagents (\textit{e.g.}, pipette), and the purpose of each action step. To avoid describing objects that are impossible to determine visually (\textit{e.g.}, the appearance of water and PBS reagent are exactly the same),  we ask the annotators to describe their visual appearance instead (\textit{e.g.}, pink reagent in a bottle with green cap). Figure~\ref{fig:cloud} shows a word cloud of the language annotations separated by views and tasks. Figure~\ref{fig:length_dist} shows the distribution of lengths of the coarse and fine level language annotations. The average lengths of the coarse and fine level annotations are 21.5 seconds and 4.6 seconds, respectively. 

\minisection{Translation \& Parsing.} For all the non-English language annotations, we translate them into English using ChatGPT. We conduct a manual check on the translation quality and use Google Translation API to translate again for unsatisfactory translations. 

\begin{figure*}
    \centering
    \includegraphics[width=\linewidth]{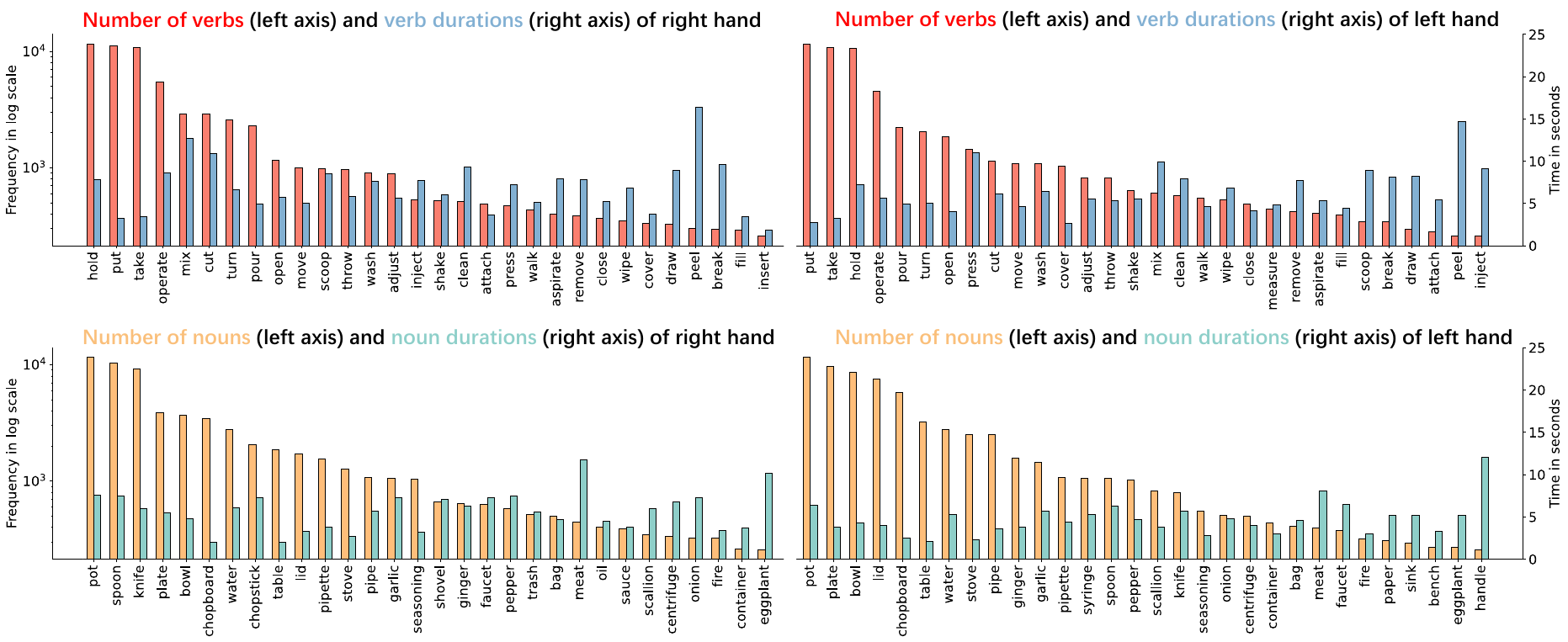}
    \caption{Occurrence and duration distribution of the annotated fine-level verbs and nouns associated with the left and right hands.}
    \label{fig:vn_complete}
\end{figure*}

To effectively parse and analyze the annotations in our dataset, we employ a rule-based framework designed to extract verbs and nouns associated with specific actions of the left and/or right hand. The process is methodical and iterative to ensure the annotation quality. The overview of the parsing is as follows:
\begin{itemize}
    \item Sentence Splitting: We begin by splitting the annotations into individual sentences using separators like commas. This step helps in isolating distinct actions or descriptions for more focused analysis.
    \item Keyword Identification and Extraction: For each split sentence, we use NLTK to identify keywords that indicate actions related to the left hand, right hand, both hands, etc. This involves analyzing the sentence structure and content to pinpoint relevant verbs and nouns. One challenge we encounter is the word ``left" itself, which can be a verb in certain contexts. To address this, we temporarily mask the mentions of left and right hands in each sentence and then re-extract the verbs and nouns. This masking helps in distinguishing between the directional use of “left” and its use as a verb.
    \item Manual Review and Iteration: After the initial extraction, we conduct a manual review of the results to identify and correct any errors. This step is crucial for ensuring the accuracy and relevance of the extracted terms. If errors are found, we revisit the first and second steps, making necessary adjustments. This iterative process continues until the manual review yields satisfactory results.
\end{itemize}

Figure~\ref{fig:vn_complete} shows the verbs and nouns extracted after associating with the left and right hands. We only show the top 30 categories due to the size limit.

\subsection{Post-processing.} 
In dealing with the practical challenges of recording egocentric videos, particularly with Pupil Invisible devices that sometimes capture footage at variable frame rates due to issues like overheating, we employ post-processing for standardization. All videos are converted to a constant frame rate of 25fps to ensure uniformity and consistency in our dataset. 

Additionally, our gaze data, which is recorded at a high frequency of 120Hz, provides detailed insights into the viewer's point of focus during the demonstration following process of the video. To ease the use of this gaze data, we align the timestamps of the egocentric camera with the eye-tracker. Once the alignment process is complete, we register each gaze data point to the temporally closest frame in the video. We then take the average of all gaze data points within one frame and use this as the final gaze data.

\begin{figure*}
    \centering
\includegraphics[width=0.98\linewidth]{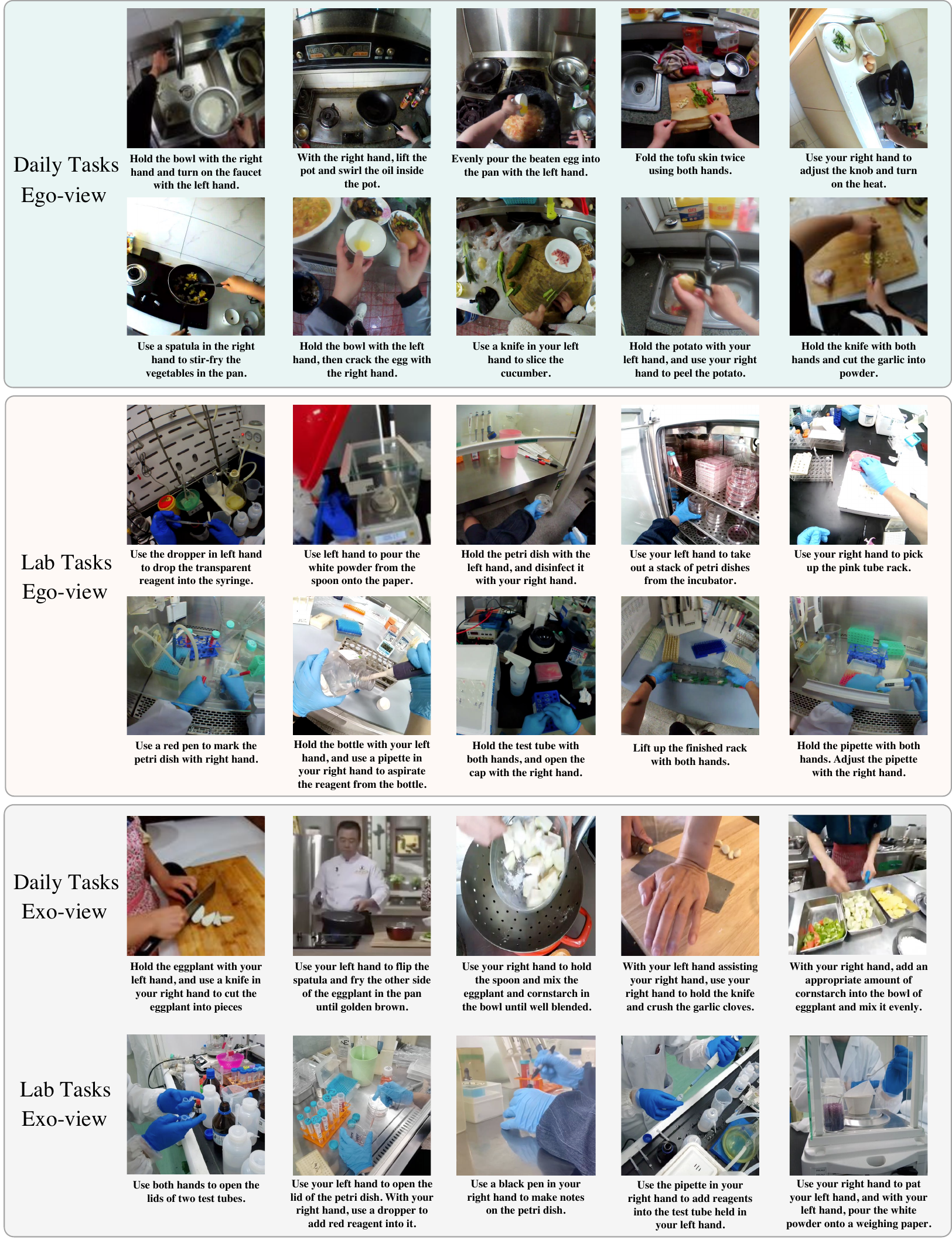}
\vspace{-0.1mm}
    \caption{Examples of video frames and corresponding fine-level language annotations in our dataset.}
    \label{fig:showdata}
\end{figure*}

In Figure~\ref{fig:showdata}, we visualize the video frames along with the annotated fine-level language annotations. \dataset features a new demonstration following setting that is a complement to existing egocentric and ego-exo datasets. Meanwhile, as can be seen in Figure~\ref{fig:showdata}, compared with existing egocentric datasets, our language annotations contain much longer sentences, enabling our dataset to be used in the captioning benchmarks.

\subsection{IRB approval}
We receive IRB approval before the data collection, adhering to the ethical standards and guidelines for research involving human participants. Participants involved in the study were provided with detailed consent forms and information sheets. These documents thoroughly explained the data capture process, the purpose of the study, and how the data would be used in the future. The consent forms, along with the information sheets, were reviewed and approved by the IRB to ensure they met all ethical standards and adequately informed participants. We maintain these documents and can provide them upon request for verification or further inquiry into our ethical and procedural practices during the data collection.

\subsection{Tasks}
Our dataset is collected for 5 types of daily tasks and three types of specialized laboratory tasks. The collection is performed in four different kitchens and three different specialized laboratories. The participants' ages range from 18 to 40 years with diverse occupations such as athletes, housekeepers, security guards, university students, and researchers. We carefully choose the daily tasks and specialized lab tasks such that a long series of procedures is needed before finishing. This can reflect the complexity of real-life activities meanwhile enabling our new benchmarks for ego-exo procedural activity bridging. Table~\ref{tab:tasks} shows the names of the 8 tasks with an example procedure. In real recordings, the procedures are usually more complicated due to repetition and other practical issues related to the environment. Note that the scientific name of the specialized reagents are not included in the language annotations but are described using their visual appearance.

\minisection{Acknowledgements.} Hosted by Shanghai AI Laboratory, Shenzhen Institutes of Advanced Technology, and Nanjing University, this work is jointly completed by a team of researchers and students from multiple institutes. Authors can be contacted via email: Yifei Huang, Mingfang Zhang, and Lijin Yang at [hyf, mfzhang, yang-lj]@iis.u-tokyo.ac.jp, Guo Chen at 602022330002@smail.nju.edu.cn. Jilan Xu can be contacted via 18210240039@fudan.edu.cn, Baoqi Pei via 12321251@zju.edu.cn, Hongjie Zhang via nju.zhanghongjie@gmail.com, and Lu Dong via dl1111@mail.ustc.edu.cn.
We give special appreciations to Yi Liu (yi.liu1@siat.ac.cn) for providing insightful ideas in the annotation phase, and Ruijie Zhang (zhangruijie@pjlab.org.cn) for the help in the arrangement of annotators.

\input{tab/tasks}

%% file: tab/cross_action_under_val.tex
\definecolor{tablegray}{gray}{0.9}
\begin{table}[t]
\centering
\scriptsize
\setlength{\tabcolsep}{5.5pt}
  \begin{tabular}[t]{lccccccc}
    \toprule
     \multirow{3}{*}{Method}&  \multirow{3}{*}{Gaze} & \multicolumn{4}{c}{Anticipation$\uparrow$} & \multicolumn{2}{c}{Planning$\downarrow$} \\
     \cmidrule(r){3-6}\cmidrule(r){7-8}
     & & Ego-V   & Ego-N & Exo-V & Exo-N & Ego & Exo  \\
    \midrule
    Exo-only & \XSolidBrush & \cellcolor{tablegray}30.7 & \cellcolor{tablegray}23.5 & \textbf{40.9} & \textbf{42.5} & \cellcolor{tablegray}83.5 & \textbf{74.6}  \\
    Ego-only & \XSolidBrush & 33.4 & 37.6 & \cellcolor{tablegray}28.7 & \cellcolor{tablegray}18.0 & 82.3 & \cellcolor{tablegray}83.7  \\
    Ego-only & \Checkmark &  \textbf{40.9} & \textbf{52.3} & \cellcolor{tablegray}37.5 & \cellcolor{tablegray}37.6  & \textbf{79.0} & \cellcolor{tablegray}81.8  \\
    Ego-only & Center & 33.4 & 38.8 & \cellcolor{tablegray}33.1 & \cellcolor{tablegray}33.7 & 81.2 & \cellcolor{tablegray}84.4  \\
    \midrule
    \multicolumn{8}{l}{\emph{Unsupervised Domain Adaption}} \\
    Ego2Exo & \XSolidBrush & 34.1 & 38.0 & \cellcolor{tablegray}34.2 & \cellcolor{tablegray}28.4 & 82.1 & \cellcolor{tablegray}83.5  \\
    Ego2Exo & \Checkmark & \textbf{41.0} & \textbf{53.7} & \cellcolor{tablegray}37.2 & \cellcolor{tablegray}37.3 & \textbf{81.5} & \cellcolor{tablegray}83.8  \\
    Exo2Ego & \XSolidBrush & \cellcolor{tablegray}31.6 & \cellcolor{tablegray}24.2 & 39.9 & \textbf{42.4} & \cellcolor{tablegray}82.9 & 77.4  \\
    Exo2Ego & \Checkmark & \cellcolor{tablegray}34.1 & \cellcolor{tablegray}31.5 & \textbf{40.2} & 42.3 & \cellcolor{tablegray}81.8 & \textbf{76.9}  \\
    \midrule
    \multicolumn{8}{l}{\emph{Knowledge Distillation}} \\
    Ego2Exo & \XSolidBrush & 30.7 & 25.1 & \cellcolor{tablegray}\textbf{41.5} & \cellcolor{tablegray}\textbf{47.6} & 83.0 & \cellcolor{tablegray}75.1 \\
    Ego2Exo & \Checkmark & 30.6 & 25.3 & \cellcolor{tablegray}41.0 & \cellcolor{tablegray}47.1 & 83.1	& \cellcolor{tablegray}\textbf{74.6} \\
    Exo2Ego & \XSolidBrush & \cellcolor{tablegray}34.6 & \cellcolor{tablegray}38.3 & 30.1 & 18.9 & \cellcolor{tablegray}81.9 & 84.9 \\
    Exo2Ego & \Checkmark& \cellcolor{tablegray}\textbf{41.2} & \cellcolor{tablegray}\textbf{55.9} & 37.0 & 39.8 & \cellcolor{tablegray}\textbf{79.0} & 82.6\\

    \midrule
    \multicolumn{8}{l}{\emph{Co-training}} \\
    Ego \& Exo & \XSolidBrush & 33.9 & 37.3 & \textbf{40.3} & 46.7 & 82.0 & 74.8  \\
    Ego \& Exo & \Checkmark & \textbf{41.6} & \textbf{52.9} & 39.6 & \textbf{47.9} & \textbf{78.3}  & \textbf{74.4} \\
    \bottomrule
  \end{tabular}
\vspace{0.5pt}
\caption{Results of cross-view action anticipation and planning benchmarks on the validation set. For anticipation, the class-mean Top-5 recall is used as the evaluation metric (higher is better). For planning, the Edit distance is used as the evaluation metric (lower is better). Gray cells show the cross-view performance.}
\vspace{-0.4cm}
\label{tab:sta-and-lta-val}
\end{table}

%% file: tab/skill_dataset_compare.tex
\begin{table}[t]
\centering
\resizebox{\columnwidth}{!}{
\begin{tabular}{lccccc}
\toprule
             & \#video clip & \#valid pairs & Av. length & Corr. exo & Gaze \\ 
 \midrule
EPIC-Skills \cite{doughty2018s}  & 216          & 2592          & 85s        & \XSolidBrush         & \XSolidBrush    \\
BEST \cite{doughty2019pros}         & 500          & 16782         & \textbf{180s}       & \XSolidBrush         & \XSolidBrush    \\
Infant Grasp \cite{li2019manipulation} & 94           & 3318          & 5s         & \XSolidBrush         & \XSolidBrush    \\
Ours         & \textbf{3304}         & \textbf{34239}         & 10s        & \Checkmark          & \Checkmark     \\ 
\bottomrule
\end{tabular}
}
\caption{Comparison of skill assessment datasets based on human pairwise ranking annotation. 
}
\label{tab:skill_datasets}
\end{table}

%% file: tab/skill_assessment_supp.tex

\begin{table}[t]
\centering
\resizebox{\columnwidth}{!}{
\begin{tabular}{lcccccc} 
\toprule
Method            & Gaze  & Egg Cracking & Peeling & Stir-fry & Cutting & Avg \\ 
\midrule
\multicolumn{6}{l}{\emph{Ego pairs only}} \\
Who's better* \cite{doughty2018s} & \XSolidBrush & 79.08 & 74.52 & 82.87 & 78.35 & 78.71    \\
RAAN* \cite{doughty2019pros}         & \XSolidBrush & 83.09 & 77.30 & 86.25 & 82.86 & 82.23    \\
Who's better* \cite{doughty2018s} & \Checkmark & 79.95 & 75.67 & 82.94 & 79.21 & 79.44    \\
RAAN* \cite{doughty2019pros}         & \Checkmark & \textbf{84.79} & 78.97 & 86.14 & 82.96 & \textbf{83.22}    \\ 
\midrule
\multicolumn{6}{l}{\emph{Ego pairs + Exo}} \\
RAAN* \cite{doughty2019pros} + RN         & \XSolidBrush & 83.14 & 77.39 & \textbf{86.47} & 82.48 & 83.01    \\
RAAN* \cite{doughty2019pros} + TL         & \XSolidBrush & 81.99 & 77.48 & 86.16 & 82.54 & 82.04    \\
RAAN* \cite{doughty2019pros} + RN         & \Checkmark & 82.84 & 78.75 & 86.19 & \textbf{83.33} & 82.78   \\
RAAN* \cite{doughty2019pros} + TL         & \Checkmark & 83.64 & \textbf{79.41} & 86.14 & 83.07 & 83.07   \\
\bottomrule
\end{tabular}
}
\vspace{-0.5pt}
\caption{Ranking accuracy of cross-view referenced skill assessment. ``*'' means using VideoMAE \cite{tong2022videomae} extracted video features. In the upper part of the table, only ego video pairs are used, while in the lower part, exo demonstrations are incorporated by ``RN'': relation network and ``TL'': triplet loss.
}
\vspace{-0.4cm}
\label{tab:skill_assess_supp}
\end{table}

%% file: tab/cross_view_tas.tex
\definecolor{tablegray}{gray}{0.9}
\begin{table*}[t]
\centering
\scriptsize
\setlength{\tabcolsep}{7.5pt}
  \begin{tabular}[t]{lccccccccccccc}
    \toprule
     \multirow{4}{*}{Method}&  \multirow{4}{*}{Gaze} & \multicolumn{6}{c}{Val} & \multicolumn{6}{c}{Test} \\
     \cmidrule(r){3-8}\cmidrule(r){9-14}
     && \multicolumn{3}{c}{Ego}  &\multicolumn{3}{c}{Exo}   &  \multicolumn{3}{c}{Ego}  &\multicolumn{3}{c}{Exo} \\
     \cmidrule(r){3-5}\cmidrule(r){6-8} \cmidrule(r){9-11}\cmidrule(r){12-14}
          && Acc & Edit & F1@Avg & Acc & Edit & F1@Avg& Acc & Edit & F1@Avg & Acc & Edit & F1@Avg\\
    \midrule
    Exo-only  & \XSolidBrush  & \cellcolor{tablegray}27.99 &\cellcolor{tablegray}33.81 & \cellcolor{tablegray}6.95 & \textbf{38.64} & \textbf{40.28} &\textbf{23.64} & \cellcolor{tablegray}24.80 & \cellcolor{tablegray}35.29 & \cellcolor{tablegray}8.13 & \textbf{42.65} & \textbf{37.32} & \textbf{20.14}\\
    Ego-only  & \XSolidBrush  & 65.35& 44.25 &40.91 & \cellcolor{tablegray}19.81 & \cellcolor{tablegray}21.18 & \cellcolor{tablegray}7.31 & 62.50 & 44.29 & 39.43  & \cellcolor{tablegray}25.09 & \cellcolor{tablegray}22.45 & \cellcolor{tablegray}6.89\\
    Ego-only & \Checkmark  &\textbf{66.01}    &\textbf{46.60} &\textbf{41.78}	& \cellcolor{tablegray}21.08 & \cellcolor{tablegray}23.29 & \cellcolor{tablegray}7.44 & \textbf{65.99} & \textbf{48.83} & \textbf{42.95} & \cellcolor{tablegray}25.14 & \cellcolor{tablegray}22.28 & \cellcolor{tablegray}8.17  \\
    Ego-only & Center  &62.14  &45.92 &36.60 & \cellcolor{tablegray}19.13 & \cellcolor{tablegray}23.02 & \cellcolor{tablegray}7.37 & 60.42 & 46.29 & 36.52 & \cellcolor{tablegray}22.83 & \cellcolor{tablegray}22.55 & \cellcolor{tablegray}7.2\\
    \midrule
    \multicolumn{10}{l}{\emph{Unsupervised Domain Adaption}} \\
    Ego2Exo & \XSolidBrush & 65.52 & 44.15 & 40.78 & \cellcolor{tablegray}20.78 & \cellcolor{tablegray}22.02 & \cellcolor{tablegray}7.55 & 63.41 & 44.35 & 40.15 & \cellcolor{tablegray}25.67 & \cellcolor{tablegray}23.12 & \cellcolor{tablegray}7.45   \\
    Ego2Exo & \Checkmark & \textbf{66.12} & \textbf{46.42} & \textbf{42.11} & \cellcolor{tablegray}21.78 & \cellcolor{tablegray}23.99 & \cellcolor{tablegray}8.43 & \textbf{65.91} & \textbf{48.81} & \textbf{42.78} & \cellcolor{tablegray}25.87 & \cellcolor{tablegray}23.44 &\cellcolor{tablegray}8.56\\
    Exo2Ego & \XSolidBrush & \cellcolor{tablegray}28.76	& \cellcolor{tablegray}33.76 & \cellcolor{tablegray}7.56 & 38.44 & 39.98 & 23.61 & \cellcolor{tablegray}25.34 & \cellcolor{tablegray}35.75 & \cellcolor{tablegray}8.67 & \textbf{42.56} & \textbf{36.71} & 20.03 \\
    Exo2Ego & \Checkmark & \cellcolor{tablegray}29.12 & \cellcolor{tablegray}34.91 & \cellcolor{tablegray}8.49 & \textbf{38.47} & \textbf{40.01} & \textbf{23.69} & \cellcolor{tablegray}27.78 & \cellcolor{tablegray}39.12 & \cellcolor{tablegray}9.87 & 42.45 & 36.66 &\textbf{21.12}\\
    \midrule
    \multicolumn{10}{l}{\emph{Knowledge Distillation}} \\
    Ego2Exo & \XSolidBrush & 33.25 & 25.68 & 9.18 & \cellcolor{tablegray}39.62 & \cellcolor{tablegray}40.36 & \cellcolor{tablegray}20.07 & 32.00 & 26.70 & 9.73 & \cellcolor{tablegray}\textbf{43.03} & \cellcolor{tablegray}38.06 & \cellcolor{tablegray}\textbf{20.44}  \\
    Ego2Exo& \Checkmark &31.16	& 28.62 & 8.60 & \cellcolor{tablegray}\textbf{40.28} &\cellcolor{tablegray}\textbf{42.24}	 & \cellcolor{tablegray}\textbf{23.09} & 29.17 & 28.49	& 8.45 & \cellcolor{tablegray}41.68 & \cellcolor{tablegray}\textbf{36.33} & \cellcolor{tablegray}20.12\\
    Exo2Ego & \XSolidBrush &\cellcolor{tablegray}65.91	& \cellcolor{tablegray}45.80 & \cellcolor{tablegray}\textbf{42.37} & 22.65  & 17.86	& 7.16 & \cellcolor{tablegray}62.77 & \cellcolor{tablegray}46.73 & \cellcolor{tablegray}41.12	& 28.20 & 17.78 & 6.06\\
    Exo2Ego & \Checkmark &\cellcolor{tablegray}\textbf{66.02}	& \cellcolor{tablegray}\textbf{47.98} & \cellcolor{tablegray}41.71 & 23.47 &24.24	 & 8.12 & \cellcolor{tablegray}\textbf{64.53} & \cellcolor{tablegray}\textbf{49.36}	& \cellcolor{tablegray}\textbf{42.24} & 28.34 & 23.49 & 7.80\\
    \midrule
    \multicolumn{10}{l}{\emph{Co-training}} \\
    Ego \& Exo & \XSolidBrush & 64.43 & 42.00 & 37.40 & 37.93 & \textbf{42.18} & \textbf{23.20} & 61.75 & 41.45 & 36.43 &41.07 & \textbf{38.73} & 21.93 \\
    Ego \& Exo & \Checkmark & \textbf{66.57} & \textbf{44.36} & \textbf{39.87} &\textbf{41.89} & 39.13 & 22.70 & \textbf{65.57} &\textbf{44.30} & \textbf{39.62} &\textbf{42.27} & 35.10 & \textbf{22.50} \\
    
    \bottomrule
  \end{tabular}
\caption{Results on cross-view temporal action segmentation benchmark. Gray cells show the cross-view performance. }
\label{tab:as}
\end{table*}

%% file: tab/cross_view_captioning.tex
    


\begin{table*}[t]
\centering
\resizebox{0.95\textwidth}{!}{
  \begin{tabular}[t]{lcccccccccc}
    \toprule
    \multirow{2}{*}{Method}& \multirow{2}{*}{Ref Train}& \multirow{2}{*}{Ref Infer}&  \multicolumn{4}{c}{Validation} & \multicolumn{4}{c}{Test} \\
    \cmidrule(r){4-7} \cmidrule(r){8-11}
    
    &&&BLEU-4 & METEOR & ROUGE-L & CIDER & BLEU-4 & METEOR & ROUGE-L & CIDER \\
    \midrule
    Exo-only &\XSolidBrush  &\XSolidBrush  & 0.024	&0.126	&0.212	&0.122 & 0.023 & 0.124	& 0.208	& 0.112\\
    Ego-only (0-shot) &\XSolidBrush  &\XSolidBrush  & 0.049	&0.116	&0.270	&0.332 & 0.048	& 0.112	& 0.266	& 0.314\\
    \midrule
    \emph{Co-training} \\
    Ego+Exo  &\Checkmark &\XSolidBrush & \textbf{0.069}&\textbf{0.139}	&\textbf{0.294}	&\textbf{0.460}  &\textbf{ 0.068} & \textbf{0.137}& \textbf{0.290} &	\textbf{0.427}\\
    \midrule
    \emph{Ref-training} \\
    Ego+Exo (1-shot)  &\Checkmark &\Checkmark & 0.047 & 0.121 &	0.275 & 0.378 &0.046 &	0.123	&0.275	&0.372 \\
    Ego+Exo (2-shot)  &\Checkmark &\Checkmark & 0.044 &	0.119 & 0.272 & 0.372 & 0.045 &	0.122 &	0.272 &	0.380\\
    \bottomrule
  \end{tabular}
}
\vspace{-0.5pt}
\caption{Cross-view referenced captioning performance. ``Ref Train/Ref Infer" refers to whether the model uses exocentric videos during training/inference.}
\vspace{-1em}
\label{tab:captioning}
\end{table*}

%% file: tab/action_recognition_zs.tex
\begin{table*}[ht]
\centering
\setlength{\tabcolsep}{5pt}
\resizebox{\linewidth}{!}{
 \begin{tabular}{lcccccccccccccccc}
 \toprule
 \multirow{5}{*}{Model} & \multicolumn{8}{c}{Val} & \multicolumn{8}{c}{Test} \\
  \cmidrule(lr){2-9} \cmidrule(lr){10-17} 
  & \multicolumn{2}{c}{Ego-Verb} & \multicolumn{2}{c}{Ego-Noun}  & \multicolumn{2}{c}{Exo-Verb} & \multicolumn{2}{c}{Exo-Noun} & \multicolumn{2}{c}{Ego-Verb} & \multicolumn{2}{c}{Ego-Noun}  & \multicolumn{2}{c}{Exo-Verb} & \multicolumn{2}{c}{Exo-Noun} \\
  \cmidrule(lr){2-3} \cmidrule(lr){4-5} \cmidrule(lr){6-7} \cmidrule(lr){8-9}  \cmidrule(lr){10-11} \cmidrule(lr){12-13} \cmidrule(lr){14-15} \cmidrule(lr){16-17} 
       & Top1  & Top5 & Top1 & Top5 & Top1  & Top5 & Top1 & Top5 & Top1  & Top5 & Top1 & Top5 & Top1  & Top5 & Top1 & Top5 \\
    \midrule 
    Oracle& 56.14 & 99.79 & 39.72  & 99.29 & 50.06 & 99.74 & 36.74 & 97.70 & 55.41 & 99.66 & 46.55 & 99.69 & 45.77 & 99.72 & 37.00& 97.59  \\
    \midrule
        CLIP~\cite{clip} & 7.89  & 22.71 & 7.08 & 19.26 & 9.49  & 22.62 & 7.70 & 20.45 & 6.96  & 21.95 & 6.39 & 18.19 & 9.02  & 20.99 & 7.09 & 19.93 \\
    \bottomrule
    \end{tabular}}
    \caption{Results of zero-shot action recognition. \textit{Oracle} denotes the upper bound of accuracy, because of the multi-label nature of clips in our dataset.}
    \label{tab:AR-ZS}

\end{table*}

%% file: tab/action_recognition_zs_prompts.tex
\begin{table}[ht]
\centering

 \begin{tabular}{ll}
 \toprule
  \# &Prompts  \\
    \midrule
   1&  A photo of action \{\}.  \\
   2&  A picture of action \{\}.  \\
   3&  Human action of \{\}.  \\
   4&  \{\}, an action.  \\
   5&  \{\} this is an action.  \\
   6&  \{\}, a video of action.  \\
   7&  Playing action of \{\}.  \\
   8& \{\} \\
   9&  Playing a kind of action, \{\}.  \\
   10&  Doing a kind of action, \{\}.  \\
   11&  Look, the human is \{\}.  \\
   12&  Can you recognize the action of \{\}?  \\
   13&  Video classification of \{\}.  \\
   14&  A video of \{\}.  \\
   15 & The man is \{\}. \\
   16 & The woman is \{\}. \\
    \bottomrule
    \end{tabular}
    \caption{Prompt templates used in the zero-shot action recognition task.}
    \label{tab:AR-ZS-prompts}

\end{table}

%% file: tab/action_recognition_ft.tex
\begin{table}[h]
\centering
\resizebox{\columnwidth}{!}{
 \begin{tabular}{llccccc}
 \toprule
  \multirow{2}{*}{View} &\multirow{2}{*}{Model}  & \multicolumn{2}{c}{Val} & \multicolumn{2}{c}{Test}  \\
  \cmidrule(lr){3-4} \cmidrule(lr){5-6} 
      & & Verb & Noun  & Verb & Noun  \\
    \midrule
      \multirow{2}{*}{Ego}&  Slowfast-R50~\cite{feichtenhofer2019slowfast} 4$\times$16 & 27.03 & 34.77 & 25.58 & 33.25 \\
      &  MViT-S~\cite{fan2021mvit} & \textbf{29.83} & \textbf{39.45} & \textbf{28.16} & \textbf{36.46} \\
    \midrule
      \multirow{2}{*}{Exo}&  Slowfast-R50~\cite{feichtenhofer2019slowfast} 4$\times$16 & 15.79 & 22.08 & 11.71 & 16.65 \\
      &  MViT-S~\cite{fan2021mvit} & \textbf{18.59} & \textbf{22.81} & \textbf{13.53} & \textbf{19.36} \\
    \bottomrule
    \end{tabular}}
    \caption{Results of fine-tuned action recognition. We utilize the multi-label mean Average Precision (mAP) evaluation metric because of the existence of multiple labels per clip. This choice is consistent with the methodology described in \cite{monfort2021mmit}. Specifically, we adopt macro mAP as the class-mean metric.} 
    \label{tab:AR-finetune}

\end{table}

%% file: tab/action_recognition_ft_param.tex
\begin{table}[t]
    \centering
    \small
    \begin{tabular}{lcc}
    \toprule
        config & Egocentric & Exocentric \\
        \midrule
        optimizer & \multicolumn{2}{c}{AdamW~\cite{kingma2014adam}} \\
        optimizer momentum & \multicolumn{2}{c}{$\beta_1, \beta_2 = 0.9, 0.999$} \\
        weight decay & \multicolumn{2}{c}{1e-4 (Slowfast), 0.05 (MViT)}  \\
        learning rate scheduler & \multicolumn{2}{c}{warmup constant} \\
        learning rate & \multicolumn{2}{c}{1e-4} \\
        batch size & 32 & 32 \\
        total epochs & 20 & 30 \\
        flip augmentation & \multicolumn{2}{c}{\checkmark} \\
        crop size & \multicolumn{2}{c}{224} \\
        randomresizedcrop & \multicolumn{2}{c}{scale=(0.08, 1)} \\
        \bottomrule
    \end{tabular}
    \caption{Training hyperparameters used for fine-tuned action recognition benchmark.}
    \label{tab:ar-ft-param}
\end{table}

%% file: tab/tasks.tex

\onecolumn
\begin{longtable}{m{0.1\linewidth}|m{0.08\linewidth}|p{0.82\linewidth}}
\toprule
Task name & Scenario & Example procedure \\
\midrule
Task1: Twice-cooked Pork & Daily &  \begin{tabular}[c]{@{}p{\linewidth}@{}}
1. Prepare spices: Take out and cut some scallion, ginger, and garlic for later use.\\
2. Prepare the pork: boil the pork together with scallion and ginger to remove impurities. Drain and set aside. Wash the pot if necessary. \\ 
3. Prepare vegetables: Take some pepper and onion, wash, and discard unused parts. \\
4. Cut vegetables: Cut the prepared vegetable into slices, and put the slices into a plate for later use. \\
5. Cut the pork: Remove the water on the pork. Use a knife to cut the pork into slices, and set aside for later use. \\
6. Stir-fry the pork: Add oil into heated pot. Then add the prepared spices into the pot. After stir-frying for about 10 seconds, add the pork into the pot. \\
7. Stir-fry the vegetables: Heat the pot and add oil, then add vegetables into the pot. Stir-fry until the vegetable is well-cooked, then add the pork into the pot.\\
8. Add seasoning: Add salt, soy sauce, sugar into the pot. Stir-fry a few times to evenly distribute the flavors. \\
9. Transfer: Transfer the cooked Twice-cooked Pork from the pot to a plate. Wash the pot if necessary.
\end{tabular} \\ \midrule
Task2: Tofu Skin with Hot Pepper & Daily &  \begin{tabular}[c]{@{}p{\linewidth}@{}}
1. Prepare tofu skin: Take out the tofu skin, fold them and cut the tofu skin into slices. \\
2. Prepare hot pepper: Take out some hot pepper, squeeze by hand and then cut into pieces. \\ 
3. Prepare spice: Take out and cut some scallion, ginger, and garlic for later use. \\
4. Boil tofu skin: Put some water into the pot, add baking soda. Boil the tofu skin until the water becomes cloudy. \\
5. Wash tofu skin:  Take out the tofu skin and put them into cold water. Wash the tofu skin such that the smell of soda diminishes. Take out and drain water.\\
6. Prepare sauce in the pot: Heat up the pot and add some oil. Put the spices into the pot. Use a spoon to put some water, soy sauce, and salt into the pot and heat up until the water boils. \\
7. Cook tofu skin: Put the tofu skin into the pot, and continue to boil until the pot becomes dry. \\
8. Cook hot pepper: Add hot pepper into the pot, stir-fry for several times. \\
9. Transfer: Add some oil into the pot, then transfer the cooked dish into a plate. \\ 
\end{tabular} \\
\midrule
Task3: Stir-fried potato, eggplant and green pepper & Daily &  \begin{tabular}[c]{@{}p{\linewidth}@{}}
1. Prepare potato: Take out some potatoes, peel and clean them. \\
2. Prepare eggplant: Take out some eggplants, remove the stems, and clean them. \\ 
3. Prepare green pepper: Take out some green pepper, remove the stems, and clean them. \\ 
4. Cut green pepper: Squeeze the green pepper using the side of the knife, then cut them into pieces.  \\ 
5. Cut eggplant: Rolling cut the eggplant into pieces. Use hand to squeeze water out of the eggplant pieces. Put some cornstarch onto the eggplant pieces and mix well. \\ 
6. Cut potato: Cut the potatoes into pieces. \\ 
7. Prepare spices: Take out and cut some scallion, ginger, and garlic for later use. \\ 
8. Prepare sauce: Take out a bowl. Add water, soy sauce, cornstarch, salt, sugar, vinegar, cooking wine into the bowl and mix them. \\ 
9. Boil potatoes: Boil some water and put the potatoes in. Take the potatoes out when the edges become transparent. \\
10. Fry vegetables: Add oil into the pot, heat the oil up and fry the peppers first and then the eggplants and then the potatoes. \\
11. Stir-fry: Add some oil into the pot, heat up and put the spices into the pot. Stir-fry a few times. Add the prepared sauce into the pot and then add all the vegetables. Stir-fry until the vegetables and the sauce are well mixed. \\
12. Transfer: Transfer the cooked dish from the pot into a plate. \\
\end{tabular} \\
\midrule
Task4: Moo Shu Pork & Daily &  \begin{tabular}[c]{@{}p{\linewidth}@{}}
1. Cut pork: Take out a piece of pork and cut into small slices. \\
2. Prepare pork: Put the pork into a bowl. Add some water and wash. Squeeze the pork and pour the water. Put the pork back and add oil, salt, and cooking wine. Mix well. \\ 
3. Prepare vegetables: Wash the necessary vegetables, use the pot to boil the vegetables. Take out for later use.\\ 
4. Prepare egg: Crack some eggs into a bowl, mix the eggs. \\ 
5. Boil vegetables: Boil some water in the pot. Add vegetables and continue to boil for a minute. \\ 
6. Fry eggs: Add oil into the pot and then fry the mixed egg. Put the fried egg scramble into a bowl. \\ 
7. Stir-fry vegetables: Heat the pot and add oil. After the oil gets heated, first add the prepared spices and then add the vegetables. Stir-fry the vegetables. \\ 
8. Stir-fry pork: Without taking the vegetables out of the pot, add pork into the pot, stir-fry all ingredients together.\\ 
9. Stir-fry egg: Without taking the ingredients out of the pot, add scrambled egg into the pot, stir-fry all ingredients together. \\ 
10. Transfer: Transfer the cooked dish from the pot into a plate. \\
\end{tabular} \\
\midrule
Task5: Tomato dough drop soup & Daily &  \begin{tabular}[c]{@{}p{\linewidth}@{}}
1. Prepare spices: Take out and cut some scallion and cumin for later use. \\
2. Prepare tomatoes: Take out tomatoes, wash and peel. \\
3. Cut tomatoes: Use a knife to cut the tomatoes first into slices and then into small pieces. \\
4. Prepare eggs: Crack eggs into a bowl, then stir until evenly mixed. \\
5. Fry eggs: Heat oil in a pan, add the evenly mixed egg mixture, and stir-fry, finally transfer the cooked scrambled eggs to a plate. \\
6. Stir-fry tomatoes: Put the chopped tomatoes into the pan, stir-fry them, and then add the scrambled eggs. \\
7. Soup-making: Add a large amount of clear water to the pot and bring it to a boil. \\
8. Prepare dough: Gradually add water to the flour while stirring until the flour forms dough. \\
9. Soup-making: Drop the flour dough into the boiling soup, while adding them, stir continuously. \\
10. Add seasoning: Add salt, pepper, and MSG (if desired) to the soup. \\
11. Transfer: Transfer the soup into a large bowl.\\
\end{tabular} \\
\midrule
Task6: Solid Phase Peptide Synthesis & Chemical lab &  \begin{tabular}[c]{@{}p{\linewidth}@{}}
1. Weighing: Use a balance to weigh the desired amount of amino acid powder (white powder). Put the powder into a test tube.\\
2. Reaction: Use a pipette to aspirate some SPPS resin (Transparent liquid) into the test tube. Shake the test tube and put the test tube onto the shaker machine. \\
3. Deprotection: Add the needed reagent into the tube to separate resin and peptide.\\
4. Suction Filteration: Take the test tube from the shaker machine, wash the peptide inside the tube, and suck the liquid into the vacuum tube. \\
5. Checking: Manual check and take necessary notes.\\
\end{tabular} \\ \midrule
Task7: Total Protein Extraction & Medical lab &  \begin{tabular}[c]{@{}p{\linewidth}@{}}
1. Preparation: Take out several test tubes, add the necessary amount of PBS reagent  (transparent liquid). Take out the cells from the fridge, disinfect, and warm the cells.\\
2. Wash cells: Use a pipette to transfer the cells into test tubes. \\
3. Centrifuge: Balance the test tubes in the centrifuge and then start the centrifugation. \\
4. Reagent making: Prepare some petri dishes, mark each dish, and add the complete medium (pink liquid) into each dish.\\
5. Transfer cells: Take the cells out of the centrifuge, and check the cell state. Transfer the cells into the prepared petri dish.\\
6. Quantification: Use an electron microscope to check the cells and record the required information. \\
7. Other necessary steps: Repeat necessary steps, make necessary reagents, etc.\\
\end{tabular} \\ \midrule
Task8: Cell subculture & Biology lab &  \begin{tabular}[c]{@{}p{\linewidth}@{}}
1. Preparation: Prepare the cell, test tubes, reagents, and petri dishes. Mark accordingly.\\
2. Wash Cells: Use a pipette to aspirate PBS reagent, use PBS to wash the cells.\\
3. Digestion: Use a separate pipette to add pancreatic enzymes (pink liquid) into the cells, put the cells into an incubator and wait for 3 minutes.\\
4. Quantification: Use an electron microscope to check the cells and record the required information.\\
5. reagent making: Prepare some petri dishes, mark each dish, and add the complete medium (pink liquid) into each dish. Use the complete medium to wash the petri dish. \\
6. Centrifuge: Balance the test tubes in the centrifuge and then start the centrifugation. \\
7. Transfer cells: Take the cells out of the centrifuge, and check the cell state. Transfer the cells into the prepared petri dish.\\
8. Incubation: Put the cells with the petri dish into the incubator.\\
\end{tabular} \\
\bottomrule
\caption{The tasks in our \dataset with example procedures. } \label{tab:tasks}
\end{longtable}
\twocolumn